%% file: aistats_2019.tex
\documentclass[twoside]{article}

\usepackage{aistats2019}
\usepackage[T1]{fontenc}    
\usepackage[hidelinks]{hyperref}       
\usepackage{url}            
\usepackage{booktabs}       
\usepackage{amsfonts}       
\usepackage{nicefrac}       
\usepackage{microtype}      

\usepackage{amsmath}
\usepackage{amssymb}
\usepackage{xcolor}
\usepackage{graphicx}
\usepackage{natbib}
\usepackage{booktabs}
\usepackage{multirow}
\usepackage{paralist}
\usepackage[export]{adjustbox}

\usepackage[textsize=scriptsize, disable]{todonotes}

\definecolor{DarkPurple}{cmyk}{0.66,1.0,0.0,0.33}
\definecolor{DarkBlue}{cmyk}{1.0,0.66,0.0,0.5}
\definecolor{DarkRed}{cmyk}{0.0,1.0,1.0,0.5}
\definecolor{DarkCyan}{cmyk}{1.0,0.0,0.1,0.5}
\definecolor{DarkGreen}{cmyk}{0.66,0.0,1.0,0.5}
\definecolor{DarkOrange}{cmyk}{0.0,0.66,1.0,0.33}

\newcommand{\cpxz}[1]{{\color{DarkCyan}{#1}}}
\newcommand{\cpx}[1]{{\color{DarkOrange}{#1}}}
\newcommand{\cpz}[1]{{\color{DarkGreen}{#1}}}

\newcommand{\cqxz}[1]{{\color{DarkPurple}{#1}}}
\newcommand{\cqx}[1]{{\color{DarkBlue}{#1}}}
\newcommand{\cqz}[1]{{\color{DarkRed}{#1}}}

\usepackage{tikz}
\newcommand*\circled[1]{\tikz[baseline=(char.base)]{
            \node[shape=circle,draw,inner sep=0.5pt] (char) {\small #1};}}
\newcommand{\term}[1]{\protect\circled{#1}}

\usepackage{array}
\newcolumntype{L}[1]{>{\raggedright\let\newline\\\arraybackslash\hspace{0pt}}m{#1}}

\usepackage{times}

\presetkeys{todonotes}{%
  backgroundcolor=blue!10!white,
  linecolor=blue!10!white,
  bordercolor=blue!10!white
}{}

\usepackage{titlesec}
\titlespacing\section{0pt}{4pt plus 2pt minus 2pt}{-2pt plus 2pt minus 0pt}
\titlespacing\subsection{0pt}{4pt plus 2pt minus 2pt}{-2pt plus 2pt minus 0pt}
\titlespacing\subsubsection{0pt}{4pt plus 2pt minus 2pt}{-2pt plus 2pt minus 0pt}

\begin{document}

%

%

\twocolumn[

\aistatstitle{Structured Disentangled Representations}

\aistatsauthor{
  Babak Esmaeili  \\
  Northeastern University\\
  \texttt{esmaeili.b@husky.neu.edu} 
  \And
  Hao Wu  \\
  Northeastern University\\
  \texttt{wu.hao10@husky.neu.edu} 
  \And
  Sarthak Jain  \\
  Northeastern University\\
  \texttt{jain.sar@husky.neu.edu} 
  \AND
  Alican Bozkurt  \\
  Northeastern University \\
  \texttt{alican@ece.neu.edu} 
  \And
  N. Siddharth  \\
  University of Oxford \\
  \texttt{nsid@robots.ox.ac.uk} 
  \And
  Brooks Paige  \\
  Alan Turing Institute \\
  University of Cambrdige \\
  \texttt{bpaige@turing.ac.uk} 
  \AND
  Dana H. Brooks \\
  Northeastern University \\
  \texttt{brooks@ece.neu.edu} 
  \And
  Jennifer Dy \\
  Northeastern University \\
  \texttt{jdy@ece.neu.edu} 
  \And
  Jan-Willem van de Meent  \\
  Northeastern University \\
  \texttt{j.vandemeent@northeastern.edu} 
  } 

\aistatsaddress{} 
]

\input{macros}

\begin{abstract}
Deep latent-variable models learn representations of high-dimensional data in an unsupervised manner.
A number of recent efforts have focused on learning representations that disentangle statistically independent axes of variation by introducing modifications to the standard objective function.
These approaches generally assume a simple diagonal Gaussian prior and as a result are not able to reliably disentangle discrete factors of variation.
We propose a two-level hierarchical objective to control relative degree of statistical independence between blocks of variables and individual variables within blocks.
We derive this objective as a generalization of the evidence lower bound, which allows us to explicitly represent the trade-offs between mutual information between data and representation, KL divergence between representation and prior, and coverage of the support of the empirical data distribution. Experiments on a variety of datasets demonstrate that our objective can not only disentangle discrete variables, but that doing so also improves disentanglement of other variables and, importantly, generalization even to unseen combinations of factors.
\end{abstract}

\section{Introduction}

Deep generative models represent data $\x$ using a low-dimensional variable $\z$ (sometimes referred to as a code). The relationship between $\x$ and $\z$ is described by a conditional probability distribution $p_\q(\x | \z)$ parameterized by a deep neural network.
There have been many recent successes in training deep generative models for complex data types such as images \citep{gatys2015neural,gulrajani2016pixelvae}, audio \citep{oord2016wavenet}, and language \citep{bowman2016generating}.
The latent code $\z$ can also serve as a compressed representation for downstream tasks such as text classification \citep{xu2017variational}, Bayesian optimization \citep{gomez2018automatic, kusner2017grammar}, and lossy image compression \citep{theis2017lossy}.
The setting in which an approximate posterior distribution $q_\f(\z | \x)$ is simultaneously learnt with the generative model via optimization of the evidence lower bound (ELBO) is known as a variational autoencoder (VAE), where $q_\f(\z | \x)$ and $p_\q(\x | \z)$ represent probabilistic encoders and decoders respectively. In contrast to VAE, inference and generative models can also be learnt jointly in an adversarial setting \citep{makhzani2015adversarial,dumoulin2016ali,donahue2016bigan}.

\begin{figure*}[!ht]
\begin{align*}
    \L(\q,\f)
    &=
    \E_{\cqxz{q_\f(\z,\x)}}
    \left[
    \log
    \frac{\cpxz{p_\q(\x,\z)}}{\cpx{p_\q(\x)} \cpz{p(\z)}}
    +
    \log
    \frac{\cqz{q_\f(\z)} \cqx{q(\x)}}{\cqxz{q_\f(\z, \x)}}
	+
    \log
    \frac{\cpx{p_\q(\x)}}{\cqx{q(\x)}}
    +
    \log \frac{\cpz{p(\z)}}{\cqz{q_\f(\z)}}
    \right],
    \\[8pt]
	&=
    \E_{\cqxz{q_\f(\z,\x)}}
    \Bigg[
    	\underbrace{%
        	\log \frac{\cpxz{p_\q(\x \mid \z)}}{\cpx{p_\q(\x)}}%
            }_{\text{\normalsize\term{1}}}
	    -
        \underbrace{%
        	\log \frac{\cqxz{q_\f(\z \mid \x)}}{\cqz{q_\f(\z)}}
            }_{\text{\normalsize\term{2}}}
    \Bigg]
    - \underbrace{%
    	\vphantom{\frac{p_\q(x)}{q_\f(z)}}%
    	\KL{\cqx{q(\x)}}{\cpx{p_\q(\x)}}}_{\text{\normalsize\term{3}}}
	- \underbrace{%
    	\vphantom{\frac{p_\q(x)}{q_\f(z)}}%
        \KL{\cqz{q_\f(\z)}}{\cpz{p(\z)}}}_{\text{\normalsize\term{4}}}.
\end{align*}
\vspace{-12pt}
\caption{\label{fig:elbo-decomp}
\emph{ELBO decomposition}. The VAE objective can be defined in terms of KL divergence between a generative model ${\cpxz{p_\q(\x,\z)}} = \cpxz{p_\q(\x \mid \z)} \cpz{p(\z)}$ and an inference model ${\cqxz{q_\f(\z,\x)}} = \cqxz{q_\f(\z \mid \x)} \cqx{q(\x)}$. We can decompose this objective into 4 terms. Term \term{1}, which can be intuitively thought of as the uniqueness of the reconstruction, is regularized by the mutual information \term{2}, which represents the uniqueness of the encoding. Minimizing the KL in term \term{3} is equivalent to maximizing the marginal likelihood $\E_{\cpx{q(\x)}}[\cqx{\log p_\q(\x)}]$. Combined maximization of $\term{1}+\term{3}$ is equivalent to maximizing $\E_{\cqxz{q_\f(\z,\x)}}[\log \cpxz{p_\q(\x \mid \z)}]$. Term \term{4} matches the inference marginal $\cqz{q_\f(\z)}$ to the prior $\cpz{p(\z)}$, which in turn ensures realistic samples $\x \sim \cpx{p_\q(\x)}$ from the generative model.
}
\vspace{-2ex}
\end{figure*}

While deep generative models often provide high-fidelity reconstructions, the representation $\z$ is generally not directly amenable to human interpretation.
In contrast to classical methods such as principal components or factor analysis, individual dimensions of $\z$ don't necessarily encode any particular semantically meaningful variation in $\x$.
This has motivated a search for ways of learning \emph{disentangled} representations, where perturbations of an individual dimension of the latent code $\z$ perturb the corresponding $\x$ in an interpretable manner.
Various strategies for weak supervision have been employed, including semi-supervision of latent variables \citep{kingma2014semi,siddharth2017learning}, triplet supervision \citep{karaletsos2015bayesian,veit2016disentangling}, or batch-level factor invariances \citep{kulkarni2015deep,bouchacourt2017multi}.
There has also been a concerted effort to develop fully unsupervised approaches that modify the VAE objective to induce disentangled representations. A well-known example  is $\beta$-VAE \citep{higgins2016beta}.
This has prompted a number of approaches that modify the VAE objective by adding, removing, or altering the weight of individual terms \citep{kumar2017variational,zhao2017infovae,gao2018auto,achille2018information}.

In this paper, we introduce hierarchically factorized VAEs (HFVAEs). The HFVAE objective is based on a two-level hierarchical decomposition of the VAE objective, which allows us to control the relative levels of statistical independence between groups of variables and for individual variables in the same group. At each level, we induce statistical independence by minimizing the \emph{total correlation} (TC), a generalization of the mutual information to more than two variables. A number of related approaches have also considered the TC \citep{kim2018disentangling, chen2018isolating, gao2018auto}, but do not employ the two-level decomposition that we consider here. In our derivation, we reinterpret the standard VAE objective as a KL divergence between a generative model and its corresponding inference model. This has the side benefit that it provides a unified perspective on trade-offs in modifications of the VAE objective.

We illustrate the power of this decomposition by disentangling discrete factors of variation from continuous variables, which remains problematic for many existing approaches. We evaluate our methodology on a variety of datasets including dSprites, MNIST, Fashion MNIST (F-MNIST), CelebA and 20NewsGroups. Inspection of the learned representations confirms that our objective uncovers interpretable features in an unsupervised setting, and quantitative metrics demonstrate improvement over related methods. Crucially, we show that the learned representations can recover combinations of latent features that were not present in any examples in the training set, which has long been an implicit goal in learning disentangled representations that is now considered explicitly.

\section{A Unified View of Generalized VAE Objectives}

\begin{figure*}[!t]
\centering
\includegraphics[width=\textwidth]{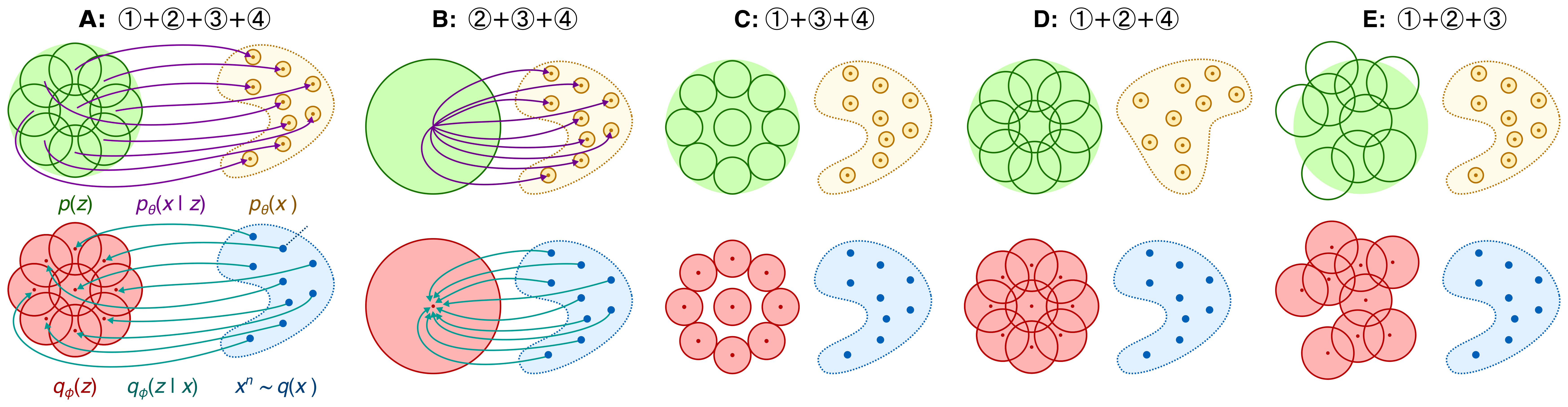}
\caption{\label{fig:vae-term-contributions} Illustration of the role of each term in the decomposition from Figure~\ref{fig:elbo-decomp}. \textbf{A:} Shows the default objective, whereas
each subsequent figure shows the effect of removing one term from the objective. \textbf{B:} Removing \term{1} means that we no longer require a unique $\z$ for each $\x^n$. Term \term{2} will then minimize $I(\x;\z)$ which means that each $\x^n$ is mapped to the prior. \textbf{C:} Removing \term{2} eliminates the constraint that $I(\x;\z)$ must be small under the inference model, causing each $\x^n$ to be mapped to a unique region in \z~space. \textbf{D:} Removing  \term{3} eliminates the constraint that $p_\q(\x)$ must match $q(\x)$. \textbf{E:} Removing \term{4} eliminates the constraint that $q_\f(\z)$ must match $p(\z)$.
}
\vspace*{-2ex}
\end{figure*}

Variational autoencoders jointly optimize two models. The generative model $p_\q(\x, \z)$ defines a distribution on a set of latent variables $\z$ and observed data $\x$ in terms of a prior $p(\z)$ and a likelihood $p_\q(\x \mid \z)$, which is often referred to as the \emph{decoder} model.
This distribution is estimated in tandem with an \emph{encoder}, a conditional
distribution $q_\f(\z \mid \x)$ that performs approximate inference in this model.
The encoder and decoder together define a probabilistic autoencoder.


The VAE objective is traditionally defined as sum over data-points $\x^n$ of the expected value of the per-datapoint ELBO, or alternatively as an expectation over an empirical distribution $q(\x)$ that approximates an unknown data distribution with a finite set of data points,
\begin{align}
	\label{eq:vae-elbo}
	\begin{split}
	    \L^{\textit{VAE}}(\q, \f)
        &:=
        \E_{q(\x)}
        \left[
        \E_{q_{\f}(\z | \x)}
        \left[
        \log
        \frac{p_{\q}(\x, \z)}{q_\f(\z \mid \x)}
        \right]
        \right]
        ,
        \\
        q(\x)
        &:=
        \frac{1}{N} \sum_{n=1}^N \delta_{\x^n}(\x)
	\end{split}
\end{align}

To better understand the various modifications of the VAE objective, which have often been introduced in an ad hoc manner, we here consider an alternate but equivalent definition of the VAE objective as a KL divergence between the generative model $p_\q(\x,\z)$ and inference model $q_\f(\z,\x) = q(\z \mid \x) q(\x)$,
\begin{align}
	\label{eq:elbo-kl}
	\L(\q, \f)
    &:=
    -\KL{q_\f(\z, \x)}{p_\q(\x,\z)}
    \\
    &=
    \E_{q_\f(\z,\x)}
    \left[
      \log
      \frac{p_\q(\x,\z)}
           {q_\f(\z \mid \x)}
    \right]
    -
    \E_{q(\x)}[\log q(\x)]. \nonumber
\end{align}
This definition differs from the expression in Equation~\eqref{eq:vae-elbo} only by a constant term $\log N$, which is the entropy of the empirical data distribution $q(\x)$. The advantage of this interpretation as a KL divergence is that it becomes more apparent what it means to optimize the objective with respect to the generative model parameters $\q$ and the inference model parameters $\f$. In particular, it is clear that the KL is minimized when $p_\q(\x,\z) = q_\f(\z,\x)$, which in turn implies that marginal distributions on data $p_\q(\x) = q(\x)$ and latent code $q_\f(\z) = p(\z)$ must also match. We will refer to $q_\f(\z)$, as the \emph{inference marginal}, which is the average over the data of the encoder distribution
\(\textstyle q_\f(\z) = \int q_\f(\z, \x) \: d\x = \frac{1}{N} \sum_{n=1}^N q_\f(\z \mid \x^n)\).

To more explicitly represent the trade-offs that are implicit in optimizing the VAE objective, we perform a decomposition (Figure~\ref{fig:elbo-decomp}) similar to the one obtained by \cite{hoffman2016elbo}. This decomposition yields 4 terms. Terms \term{3} and \term{4} enforce consistency between the marginal distributions over $\x$ and $\z$. Minimizing the KL in term \term{3} maximizes the marginal likelihood $\E_{q(\x)}[\log p_\q(\x)]$, whereas minimizing \term{4} ensures that the inference marginal $q_\f(\z)$ approximates the prior $p(\z)$. 
Terms \term{1} and \term{2} enforce consistency between the conditional distributions. Intuitively speaking, term \term{1} maximizes the identifiability of the values $\z$ that generate each $\x^n$; when we sample $\z \sim q_\f(\z \mid \x^n)$, then the likelihood $p_\q(\x^n \mid \z)$ under the generative model should be \emph{higher} than the marginal likelihood $p_\q(\x^n)$. Term \term{2} regularizes term \term{1} by minimizing the mutual information $I(\z;\x)$ in the inference model, which means that $q_\f(\z \mid \x^n)$ maps each $\x^n$ to less identifiable values.

Note that term \term{1} is intractable in practice, since we are not able to pointwise evaluate $p_\q(\x)$. We can circumvent this intractability by combining \term{1} + \term{3} into a single term, which recovers the likelihood 
\begin{align*}
	&
	\argmax_{\q,\f}
	\E_{q_\f(\z,\x)}
    \left[
    	\log
        \frac{p_\q(\x \mid \z)}
             {p_\q(\x)}
	+
    \log \frac{p_\q(\x)}{q(\x)}
    \right]
    \\
    &=
	\argmax_{\q,\f}
	\E_{q_\f(\z,\x)}
    \left[
    	\log
        p_\q(\x \mid \z)
    \right].
\end{align*}

To build intuition for the impact of each of these terms, Figure~\ref{fig:vae-term-contributions} shows the effect of removing each term from the objective. When we remove \term{3} or \term{4} we can learn models in which $p_\q(\x)$ deviates from $q(\x)$, or $q_\f(\z)$ deviates from $p(\z)$. When we remove \term{1}, we eliminate the requirement that $p_\q(\x^n \mid \z)$ should be higher when $\z \sim q_\f(\z \mid \x^n)$ than when $\z \sim p(\z)$. Provided the decoder model is sufficiently expressive, we would then learn a generative model that ignores the latent code $\z$. This undesirable type of solution does in fact arise in certain cases, even when \term{1} is included in the objective, particularly when using auto-regressive decoder architectures \citep{chen2016variational}.

\begin{figure*}[!t]
\begin{align*}
    \small
    -\KL{\cqz{q_\f(\z)}}{\cpz{p(\z)}}
    &=
    -
    \E_{\cqz{q_\f(\z)}}
    \left[
    \log
    \frac{\cqz{q_\f(\z)}}{\prod_d \cqz{q_\f(\z_d)}}
    +
	\log
	\frac{\prod_d \cqz{q_\f(\z_d)}}{\prod_{d} \cpz{p(\z_d)}}
    +
	\log
	\frac{\prod_{d} \cpz{p(\z_d)}}{\cpz{p(\z)}}
    \right]
    \\[8pt]
    &=
    \phantom{-}
    \E_{\cqz{q_\f(\z)}}
    \underbrace{%
	  \left[
	  \log
	  \frac{\cpz{p(\z)}}{\prod_{d} \cpz{p(\z_d)}}
	  -
       \log
      \frac{\cqz{q_\f(\z)}}{\prod_d \cqz{q_\f(\z_d)}}
      \right]
    }_{\text{\normalsize\term{\textsc{a}}}}
    -
    \sum_d
	\underbrace{%
    	\vphantom{\frac{p_\q(x)}{q_\f(z)}}
    	\KL{\cqz{q_\f(\z_d)}}{\cpz{p(\z_d)}}
    }_{\text{\normalsize\term{\textsc{b}}}}.
    \\
    \term{\textsc{b}} 
    =
    \phantom{-}
    \E_{\cqz{q_\f(\z_d)}}
    &\underbrace{%
      \left[
  	  \log
	  \frac{\cpz{p(\z_d)}}{\prod_{e} \cpz{p(\z_{d,e})}}
      -
	  \log
      \frac{\cqz{q_\f(\z_d)}}{\prod_e \cqz{q_\f(\z_{d,e})}}
      \right]
    }_{\text{\normalsize\term{i}}}
    -
    \sum_e
    \underbrace{%
    	\vphantom{\frac{p_\q(x)}{q_\f(z)}}
		\KL{\cqz{q_\f(\z_{d,e})}}{\cpz{p(\z_{d,e})}}
    }_{\text{\normalsize\term{ii}}}
\end{align*}
\vspace{-12pt}
\caption{%
  \label{fig:kl-decomp}
  \emph{Hierarchical KL decomposition}. We can decompose term \term{4} into subcomponents \term{\textsc{a}} and \term{\textsc{b}}. Term \term{\textsc{a}} matches the total correlation between variables in the inference model relative to the total correlation under the generative model. Term \term{\textsc{b}} minimizes the KL divergence between the inference marginal and prior marginal for each variable $\z_d$. When the variable $\z_d$ contains sub-variables $\z_{d,e}$, we can recursively decompose the KL on the marginals $\z_d$ into term \term{i}, which matches the total correlation, and term \term{ii}, which minimizes the per-dimension KL divergence.}
\vspace*{-2ex}
\end{figure*}

When we remove \term{2}, we learn a model that minimizes the overlap between $q_\f(\z \mid \x^n)$ for different data points $\x^n$, in order to maximize \term{1}. This maximizes the mutual information $I(\x;\z)$, which is upper-bounded by $\log N$. In practice \term{2} often saturates to $\log N$, even when included in the objective, which suggests that maximizing \term{1} outweighs this cost, at least for the encoder/decoder architectures that are commonly considered in present-day models.

\section{Hierarchically Factorized VAEs (HFVAEs)}

In this paper, we are interested in defining an objective that will encourage statistical independence between features. The $\beta$-VAE objective \citep{higgins2016beta} aims to achieve this goal by defining the objective
\begin{align*}
	\L^{\beta\text{-VAE}}(\q,\f)
    =
	\E_{q(\x)}
      \Big[
      	&
	    \E_{q_\f(\z | \x)}
        \left[
          \log p_\q(\x | \z)
        \right]
        \\
        &
        -
    	\beta \KL{q_\f(\z | \x)}{p(\z)}
    \Big].
\end{align*}

We can express this objective in the terms of Figure~\ref{fig:elbo-decomp} as $\term{1} + \term{3} + \beta \: \left( \term{2} + \term{4} \right)$.
In order to induce disentangled representations, the authors set $\beta > 1$. This works well in certain cases, but it has the drawback that it also increases the strength of \term{2}, which means that the encoder model may discard more information about $\x$ in order to minimize the mutual information $I(\x;\z)$.


Looking at the $\beta$-VAE objective, it seems intuitive that increasing the weight of term \term{4} is likely to aid disentanglement. One notion of disentanglement is that there should be a low degree of correlation between  different latent variables $\z_d$. If we choose a mean-field prior $p(\z) = \prod_d p(\z_d)$, then minimizing the KL term should induce an inference marginal $q_\f(\z) = \prod_d q_\f(\z_d)$ in which $\z_d$ are also independent. However, in addition to being sensitive to correlations, the KL will also be sensitive to discrepancies in the shape of the distribution. When our primary interest is to disentangle representations, then we may wish to relax the constraint that the shape of the distribution matches the prior in favor of enforcing statistical independence.

To make this intuition explicit, we decompose \term{4} into two terms \term{\textsc{a}} and \term{\textsc{b}} (Figure~\ref{fig:kl-decomp}). As with term \term{1} + \term{2}, term \term{\textsc{a}} consists of two components. The second of these takes the form of a total correlation, which is the generalization of the mutual information to more than two variables,
\begin{align}
    \begin{split}
        TC(\z)
        &=
        \E_{q_\f(\z)}
        \left[
        	\log \frac{q_\f(\z)}{\prod_d q_\f(\z_d)}
        \right]
        \\
        &=
        \text{KL}
        \Big(
            q_\f(\z)
            \Big|\Big|
            \textstyle
            \prod_d q_\f(\z_d)
        \Big)
    \end{split}
\end{align}
Minimizing the total correlation yields a $q_\f(\z)$ in which different $\z_d$ are statistically independent, hereby providing a possible mechanism for inducing disentanglement. In cases where $\z_d$ itself represents a group of variables, rather than a single variable, we can continue to decompose to another set of terms \term{i} and \term{ii} which match the total correlation for $\z_d$ and the KL divergences for constituent variables $\z_{d,e}$. This provides an opportunity to induce hierarchies of disentangled features. We can in principle continue this decomposition for any number of levels to define an HFVAE objective. We here restrict ourselves to the two-level case, which corresponds to an objective of the form
\begin{align}
	\begin{split}
	\L^{\text{HFVAE}}(\q, \f)
    &=
    \term{1}+\term{3}+\term{ii}
    + \alpha \: \term{2}
    + \beta \: \term{\textsc{a}}
    + \gamma \: \term{i}.
    \end{split}
\end{align}
In this objective, $\alpha$ controls the $I(\x;\z)$ regularization, $\beta$ controls the TC regularization between groups of variables, and $\gamma$ controls the TC regularization within groups.
This objective is similar to, but more general than, the one recently proposed by \citet{kim2018disentangling} and \citet{chen2018isolating}. Our objective admits these objectives as a special case corresponding to a non-hierarchical decomposition in which $\beta = \gamma$. The first component of \term{\textsc{a}} is not present in these objectives, which implicitly assume that $p(\z) = \prod_d p(\z_d)$.
In the more general case where $p(\z) \not= \prod_d p(\z_d)$, maximizing \term{\textsc{a}} with respect to $\phi$ will match the total correlation in $q(\z)$ to that in $p(\z)$.


\begin{figure*}[t!]
  \centering
  \begin{tabular}{@{}c@{\hspace*{2pt}}c@{\hspace*{4pt}}c@{\hspace*{4pt}}c@{}}
    & \textsf{Orientation}
    & \textsf{Smiling}
    & \textsf{Sunglasses} \\[1ex]
    \rotatebox[origin=lt]{90}{\quad\textsf{HFVAE}}
    & \includegraphics[width=0.32\linewidth, height=0.1\textheight]{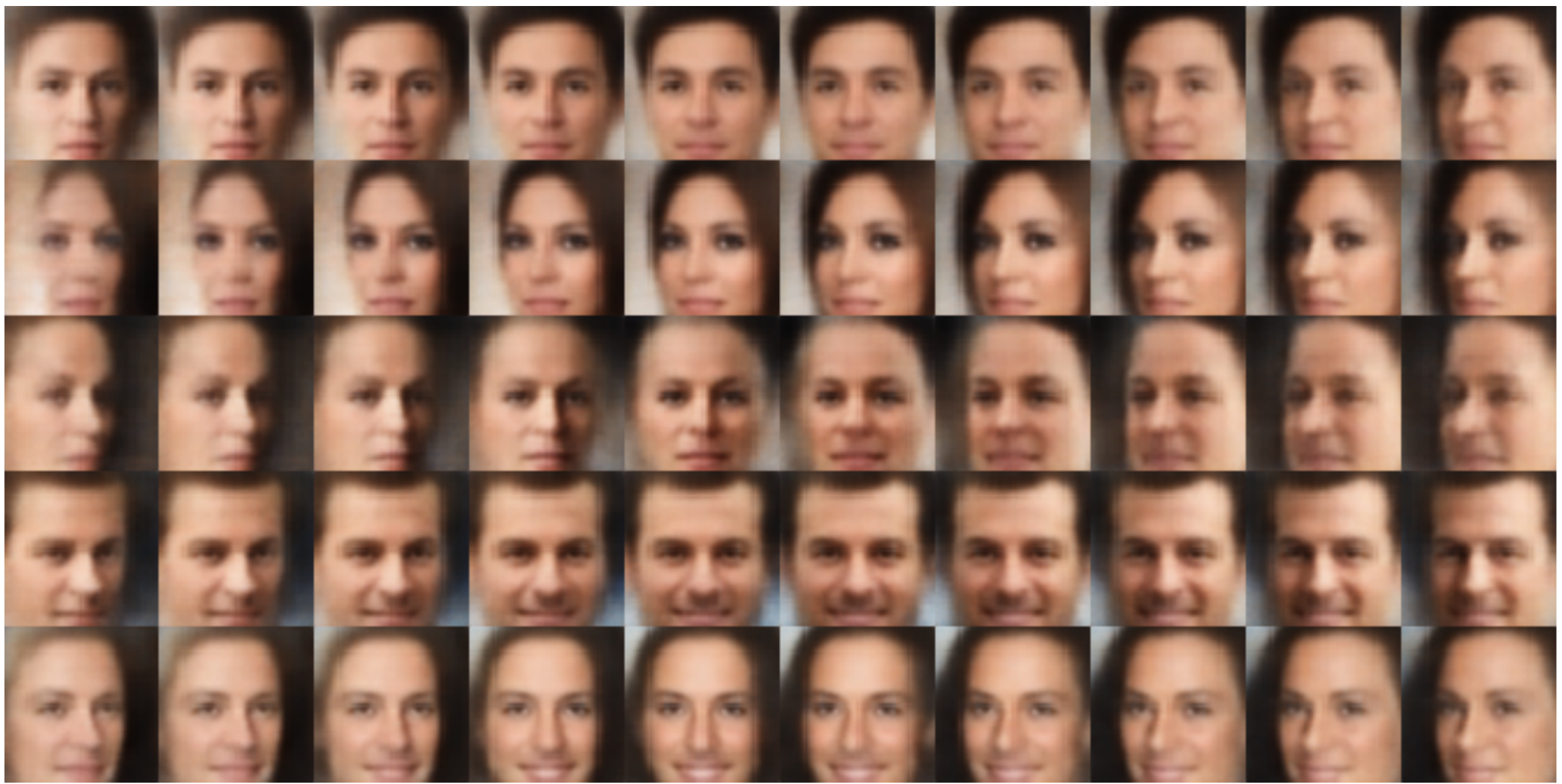}
    & \includegraphics[width=0.32\linewidth, height=0.1\textheight]{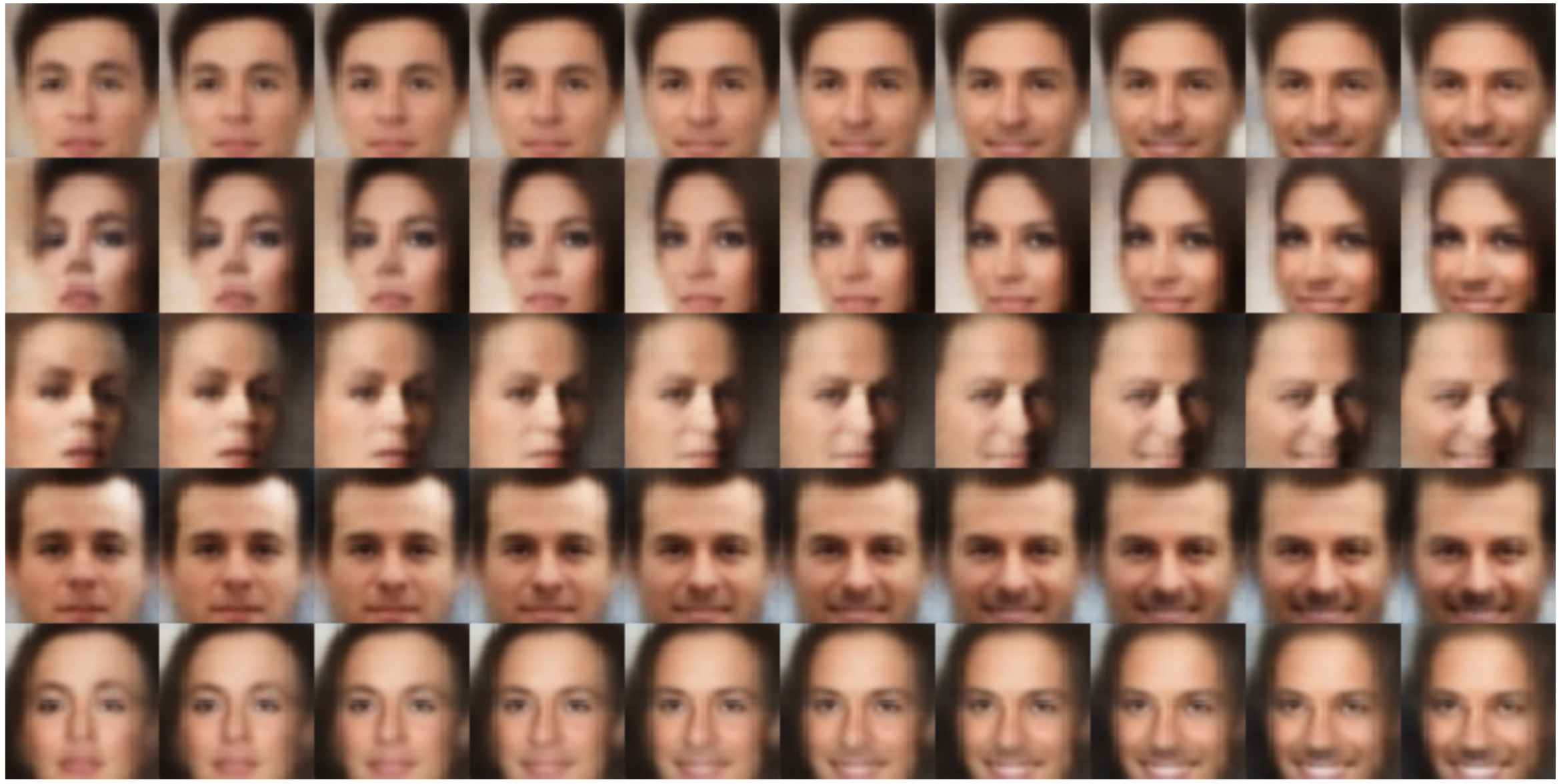}
    & \includegraphics[width=0.32\linewidth, height=0.1\textheight]{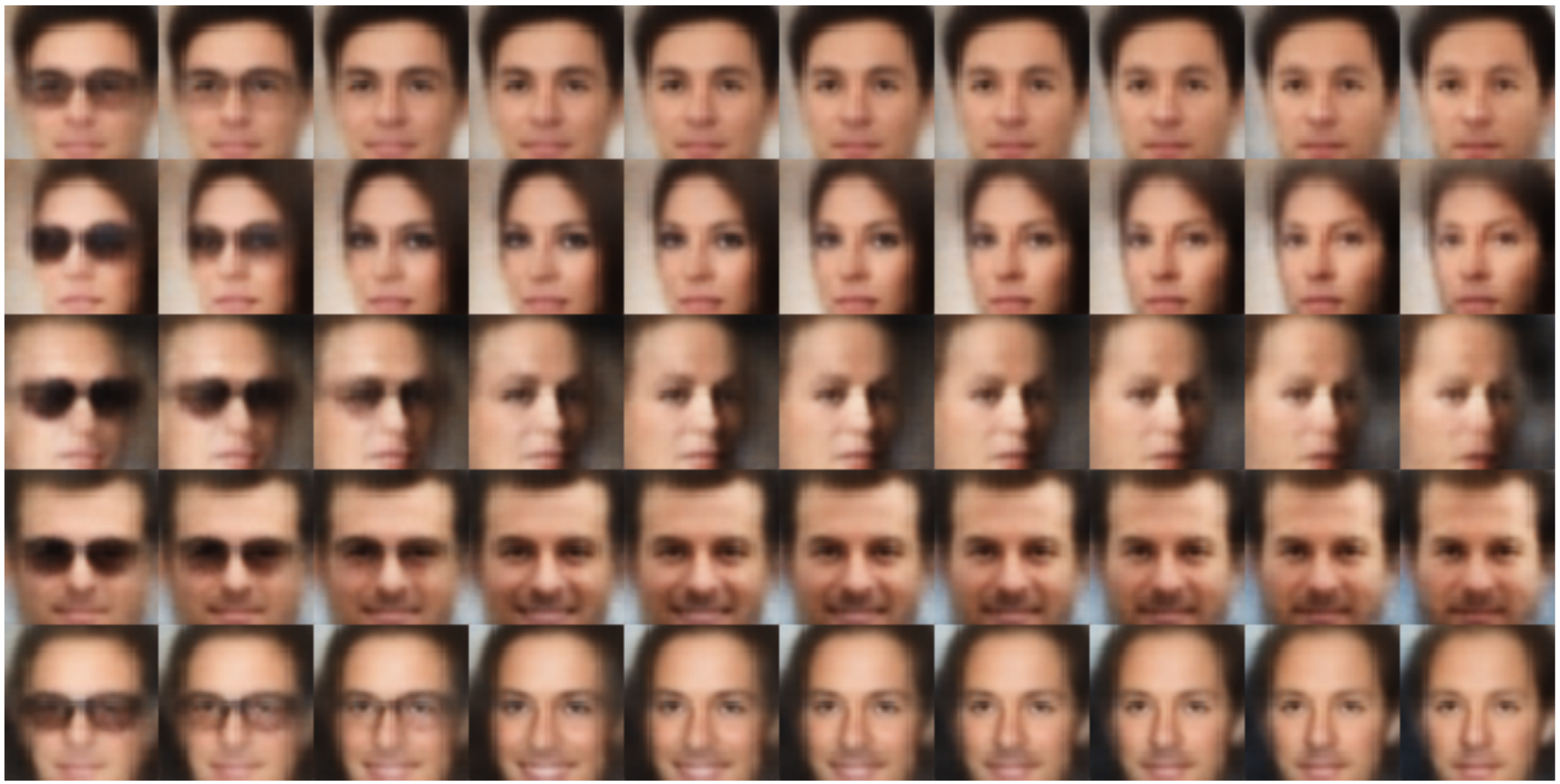} \\[0.7ex]
    \rotatebox[origin=lt]{90}{\quad$\beta$-\textsf{VAE}}
    & \includegraphics[width=0.32\linewidth, height=0.1\textheight]{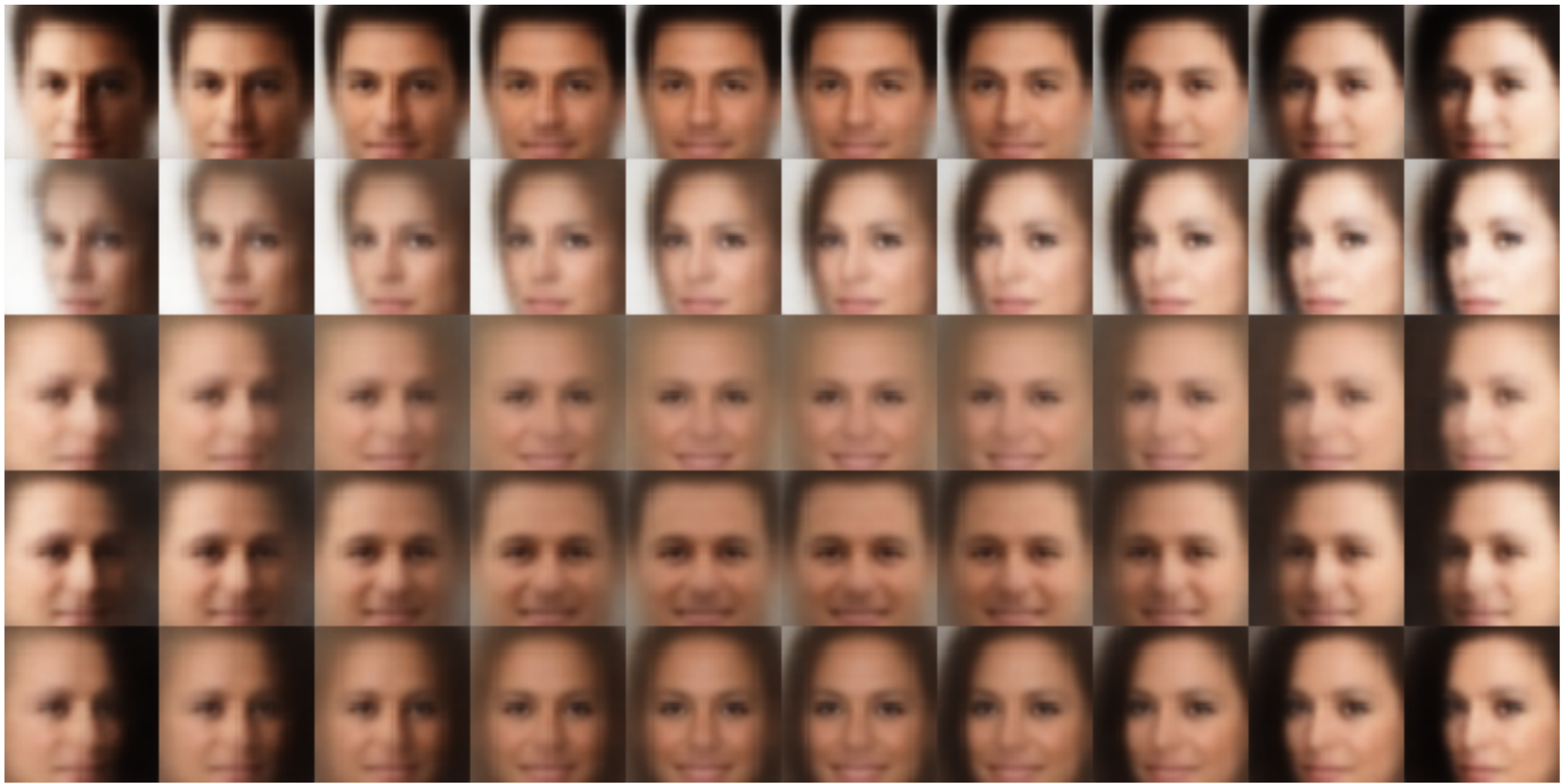}
    & \includegraphics[width=0.32\linewidth, height=0.1\textheight]{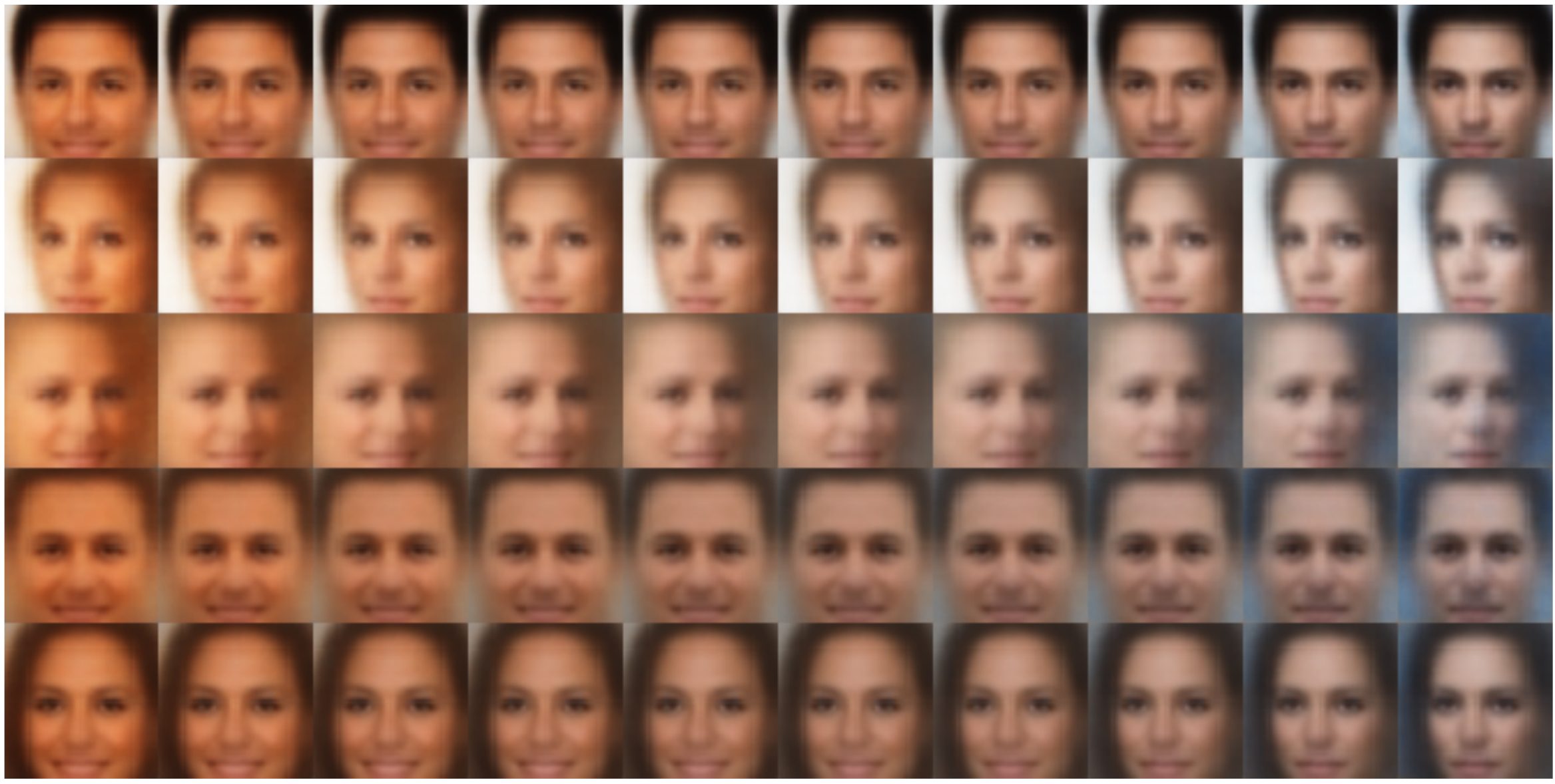}
    & \includegraphics[width=0.32\linewidth, height=0.1\textheight]{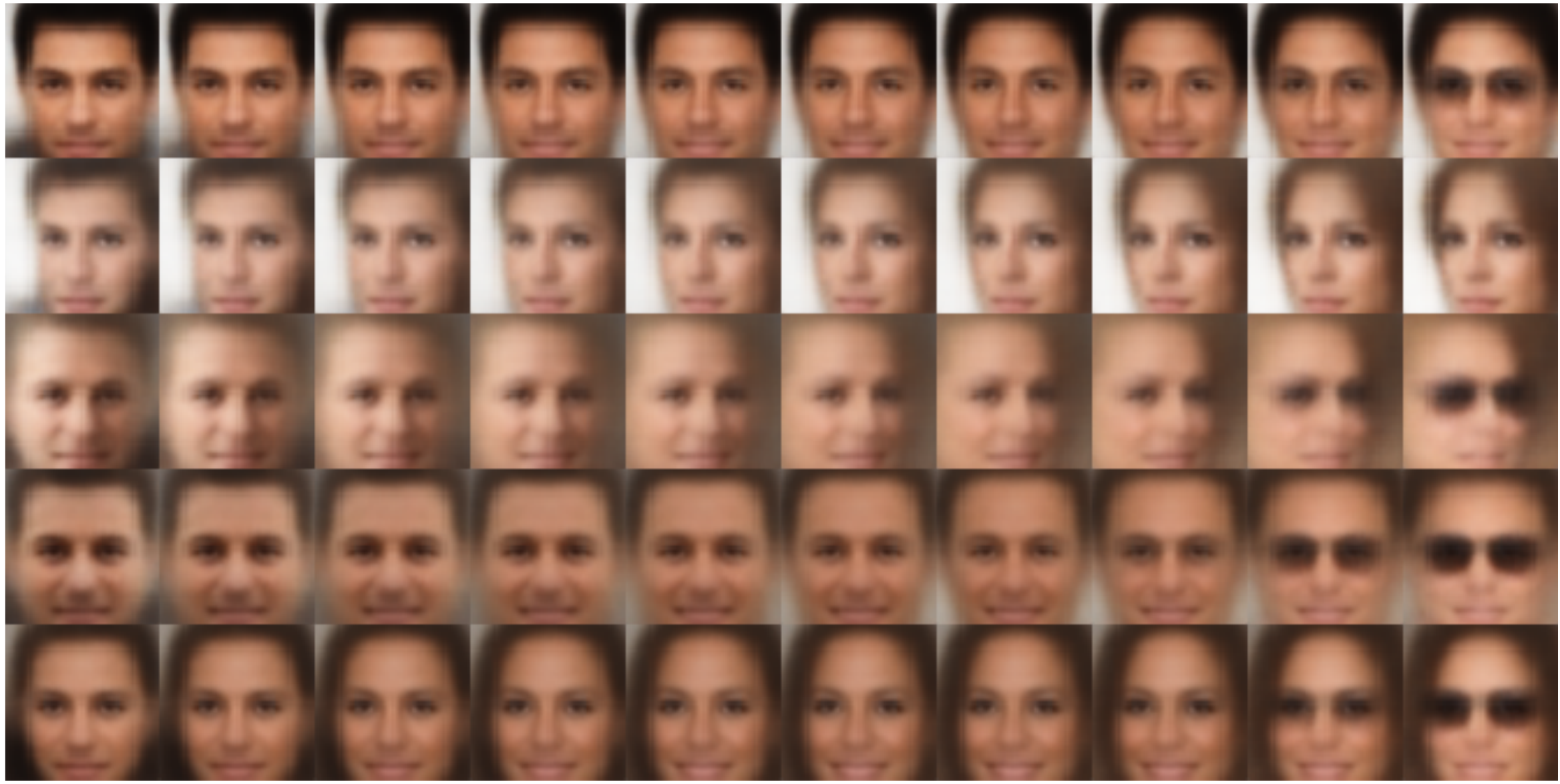}
  \end{tabular}
  \caption{Interpretable factors in CelebA for a HFVAE ($\beta=5.0$, $\gamma=3.0$) and a $\beta$-VAE ($\beta=8.0$)}
  \label{fig:CelebA}
  \vspace*{-1ex}
\end{figure*}



\subsection{Approximation of the Objective}

In order to optimize this objective, we need to approximate the inference marginals $q_\f(\z)$, $q_\f(\z_d)$, and $q_\f(\z_{d,e})$. Computing these quantities exactly requires a full pass over the dataset, since $q_\f(\z)$ is a mixture over all data points in the training set. We approximate $q_\f(\z)$ with a Monte Carlo estimate $\hat{q}_\f(\z)$ over the same batch of samples that we use to approximate all other terms in the objective $\L^{\text{HFVAE}}(\q,\f)$. For simplicity we will consider the term
\begin{align}
	\label{eq:inf-marginal-mc}
	&
	\E_{q_\f(\z,\x)}[\log q_\f(\z)]
    \simeq
    \frac{1}{B}
    \sum_{b=1}^B
    \log q_\f(\z^{b}),
    \\
    &
    \quad\z^{b} \sim q_\f(\z \mid \x^{b}),
    \quad
    b \sim \text{Uniform}(1, \ldots, N) \nonumber
\end{align}
We define the estimate of $\hat{q}_\f(\z^{b})$ as (see Appendix~\ref{sec:appendix-qz})
\begin{align}
    \begin{split}
        \hat{q}_\f(\z^{b})
        &:=
        \frac{1}{N}
        q_\f(\z^{b} \mid \x^{b})
        +
        \frac{N-1}{N (B-1)}
        \sum_{b' \not= b}
        q_\f(\z^{b} \mid \x^{b'}).
    \end{split}
\end{align}
This estimator differs from the one in \citet{kim2018disentangling}, which is based on adversarial-style estimation of the density ratio, and is also distinct from the estimators in \citet{chen2018isolating} who employ different approximations. We can think of this approximation as a partially stratified sample, in which we deterministically include the term $\x^n = \x^{b}$ and compute a Monte Carlo estimate over the remaining terms, treating indices $b'\not=b$ as samples from the distribution $q(\x \mid \x \not= \x^{b})$.  We now substitute $\log \hat{q}_\f(\z)$ for $\log q_\f(\z)$ in Equation~\eqref{eq:inf-marginal-mc}. By Jensen's inequality this yields a lower bound on the original expectation. While this induces a bias, the estimator is consistent. In practice, the bias is likely to be small given the batch sizes (512-1024) needed to approximate the inference marginal.

\section{Related Work}

In addition to the work of \cite{kim2018disentangling} and \cite{chen2018isolating}, our objective is also related to, and generalizes, a number of recently-proposed modifications to the VAE objective (see Table~\ref{tab:comparison} for an overview). \citet{zhao2017infovae} considers an objective that eliminates the mutual information in \term{2} entirely and assigns an additional weight to the KL divergence in \term{4}. \citet{kumar2017variational} approximate the KL divergence in \term{4} by matching the covariance of $q_\f(\z)$ and $p(\z)$. Recent work by \citet{gao2018auto} connects VAEs to the principle of correlation explanation, and defines an objective that reduces the mutual information regularization in \term{2} for a subset of ``Anchor'' variables $\z_a$. \citet{achille2018information} interpret VAEs from an information-bottleneck perspective and introduce an additional TC term into the objective. In addition to VAEs, generative adversarial networks (GANs) have also been used to learn disentangled representations. The InfoGAN \citep{chen2016infogan} achieves disentanglement by \emph{maximizing} the mutual information between individual features and the data under the \emph{generative model}.  

In settings where we are not primarily interested in inducing disentangled representations, the $\beta$-VAE objective has also been used with $\beta < 1$ in order to improve the quality of reconstructions \citep{alemi2016deep,engel2017latent,liang2018variational}. While this also decreases the relative weight of \term{2}, in practice it does not influence the learned representation in cases where $I(\x;\z)$ saturates anyway. The tradeoff between likelihood and the KL term and the influence of penalizing the KL term on mutual information has been studied more in depth in \cite{alemi2018fixing, burgess2018understanding}. In other recent work, \cite{dupont2018joint} considered models containing both Concrete and Gaussian variables. However, the objective was not decomposed to get \term{\textsc{a}}, but was based on the objective proposed by \cite{burgess2018understanding}.

\begin{table}[tb]
  \centering
  \resizebox{\columnwidth}{!}{%
  \renewcommand{\arraystretch}{1.3}
  \begin{tabular}{L{4cm}l@{\hspace*{10pt}}l@{}}
    \toprule
    \textbf{Paper}
    & \textbf{Objective} \\
    \midrule
    \citet{kingma2013auto-encoding}, \newline\citet{rezende2014stochastic}
    & $\term{1}+\term{2}+\term{3}+\term{4}$ \\[0.5ex]
    \citet{higgins2016beta}
    & $\term{1} + \term{3} + \beta \left(\term{2} + \term{4}\right)$ \\[0.5ex]
    \citet{kumar2017variational}
    & $\term{1} + \term{2} + \term{3} + \lambda\, \term{4}$ \\[0.5ex]
    \citet{zhao2017infovae}
    & $\term{1} + \term{3} + \lambda\, \term{4}$ \\[0.5ex]
    \citet{alemi2018fixing}, \newline\citet{burgess2018understanding}
    & $\term{1} + \term{3} + \gamma|(\term{2} + \term{4}) - C|$\\[0.5ex]
    \citet{gao2018auto}
    & $\term{1} + \term{2} + \term{3} + \term{4} - \lambda \: \term{2}^\text{a}$	\\[0.5ex]
    \cite{achille2018information}
    & $\term{1} + \term{3} + \beta\, \term{2} + \gamma\, \term{\textsc{a}}^*$ \\[0.5ex]
	\citet{kim2018disentangling}, \newline\citet{chen2018isolating}
    & $\term{1} + \term{2} + \term{3} + \term{\textsc{b}} + \beta\, \term{\textsc{a}}^*$ \\[0.5ex]
    HFVAE (this paper)
    & $\term{1} + \term{3} + \term{ii} + \alpha\, \term{2} + \beta\, \term{\textsc{a}} + \gamma\, \term{i}$ \\
    \bottomrule
  \end{tabular}}
  \vspace{-1em}


  \caption{Comparison of objectives in autoencoding deep generative models. The asterisk $\term{\textsc{a}}^*$ indicates that the prior factorizes, i.e. $p(\z)=\prod_d p(\z_d)$. The notation $\term{2}^\text{a}$ refers to restriction of the mutual information \term{2} to a subset of "Anchor" variables $\z_a$.}
  \label{tab:comparison}
\end{table}

\section{Experiments}

\begin{figure*}[!t]
  \centering
  \adjustbox{valign=t}{%
  \includegraphics[height=3cm]{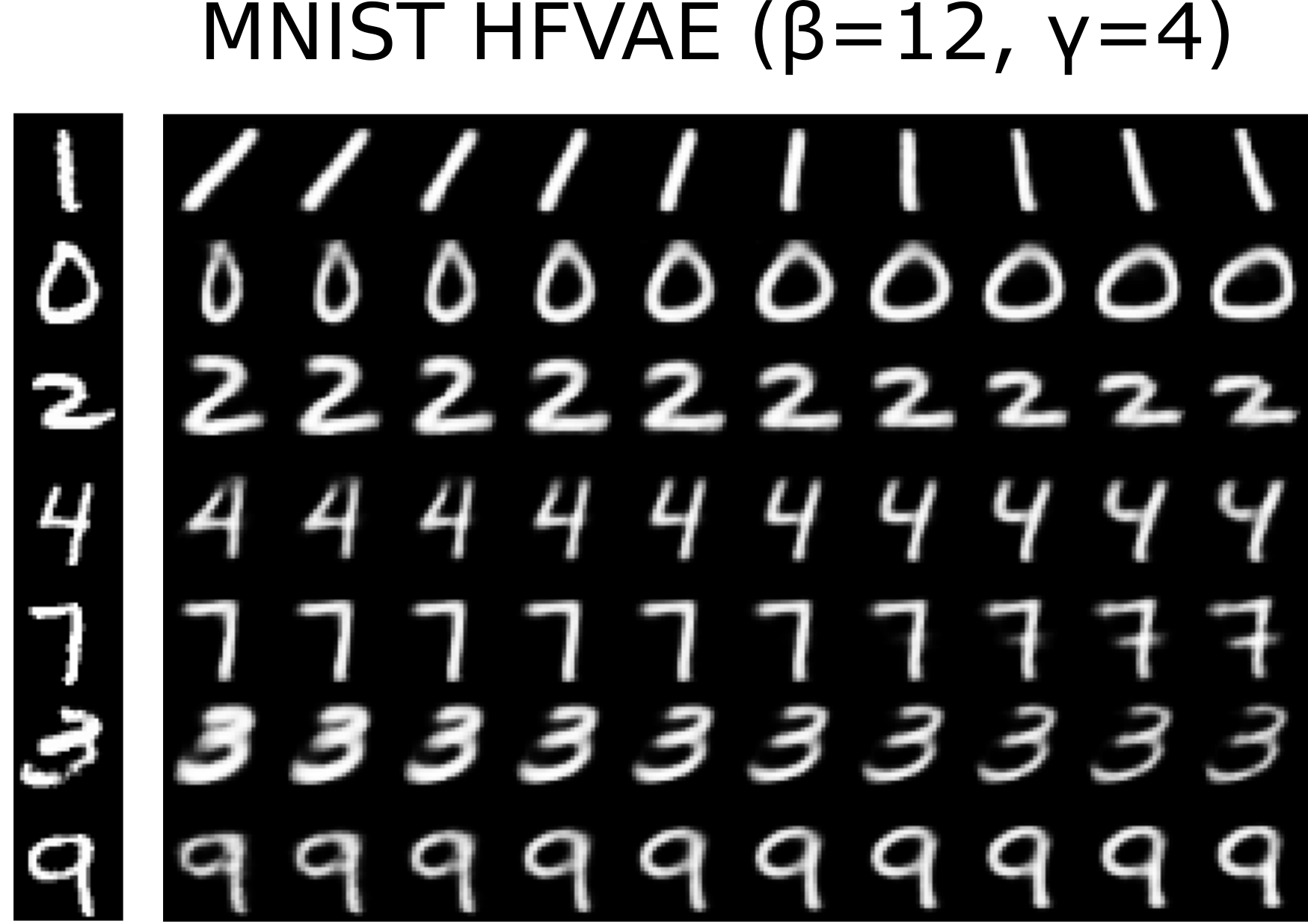}}
  \hspace{0.01\textwidth}
  \adjustbox{valign=t}{%
  \includegraphics[height=3cm]{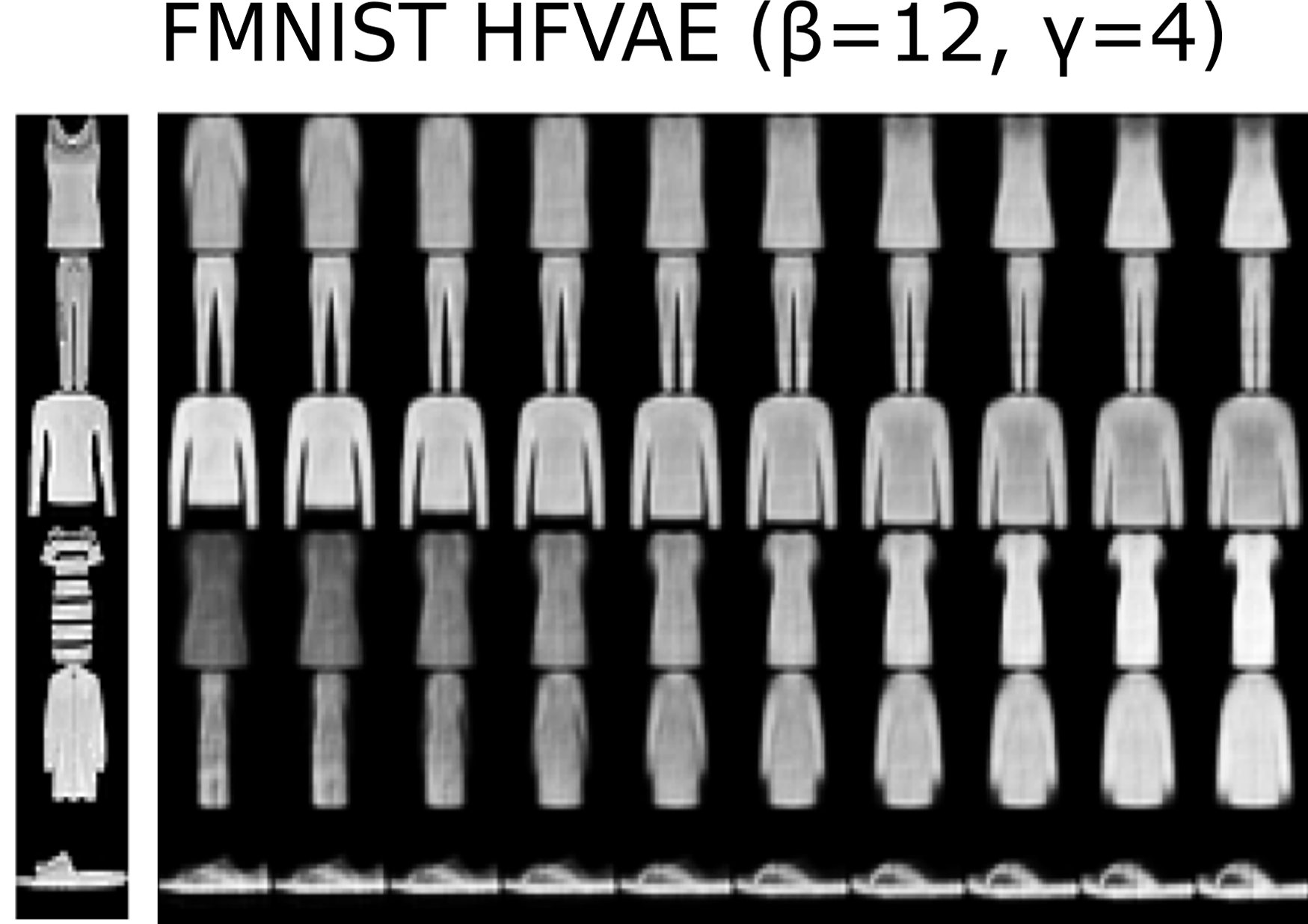}}
  \hspace{0.01\textwidth}
  \adjustbox{valign=t}{%
     \includegraphics[height=3.4cm]{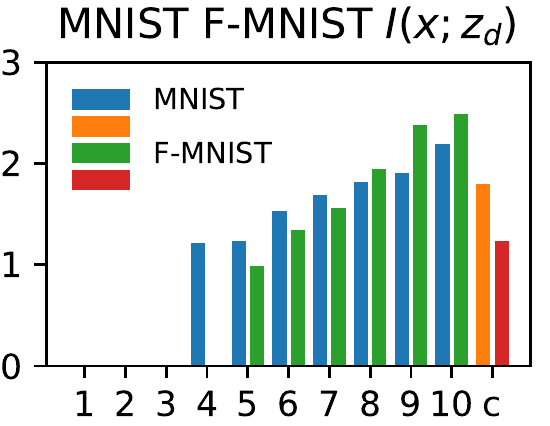}}
  \vspace*{-2ex}
  \caption{Left: MNIST and F-MNIST reconstructions for $\z_d \in (-3, 3)$. Rows contain both different samples from data and different dimensions $d$. Right: The mutual information \term{2} for each individual dimension $I(\x; \z_{d})$, ranked in ascending order, with the Concrete variable shown last. The HFVAE \emph{prunes} 3 continuous dimensions in MNIST and 4 in F-MNIST.}
  \label{fig:mnist-fmnist}
\end{figure*}

\begin{figure*}[!t]
  \centering
  \includegraphics[width=\textwidth]{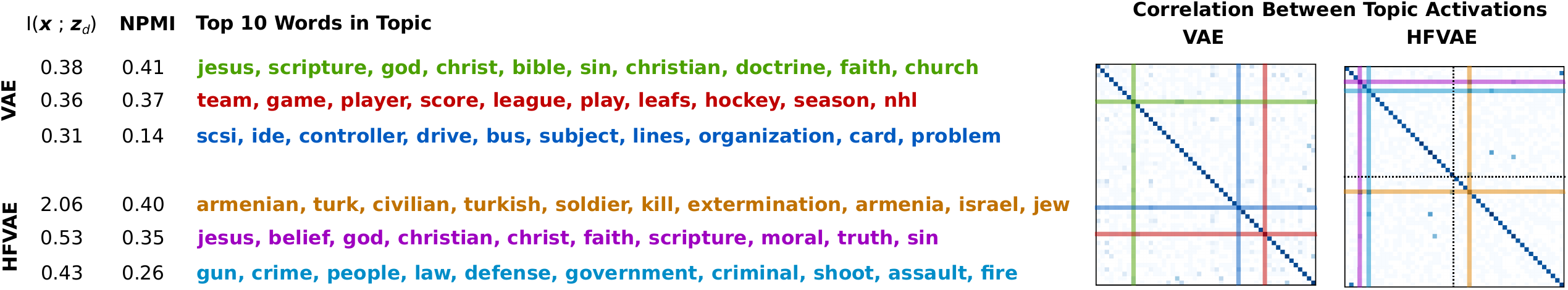}
  \vspace*{-2ex}
  \caption{Learned topics in the 20NewsGroups dataset using the HFVAE and the VAE objective. The middle column shows frequent words for the 3 most informative topics of the VAE, and the 3 \emph{most correlated} topics in the HFVAE\@. The left column lists their corresponding mutual information with $\x$ and the topic coherence score. The right column shows the correlations between topics. The HFVAE learns 2 groups of topics that are internally uncorrelated (top-left and bottom-right quadrants), whilst uncovering sparse correlations between groups (top-right and bottom-left quadrants).}
  \label{fig:prod-lda}
\end{figure*}

To assess the quality of disentangled representations that the HFVAE induces, we evaluate a number of tasks and datasets. We consider \textit{CelebA} \citep{liu2015deep} and \textit{dSprites} \citep{higgins2016beta} as exemplars of datasets that are typically used to demonstrate general-purpose disentangling. As specific examples of datasets that require a discrete variable, we consider \textit{MNIST} \citep{lecun2010mnist} and \textit{F-MNIST} \citep{xiao2017fashion}. Finally, we consider an example that extends beyond image-based domains by using the HFVAE objective to train neural topic models on the \textit{20NewsGroups} \citep{20newsgroups} dataset.
We compare a number of objectives and priors on these datasets, including the standard VAE objective \citep{kingma2013auto-encoding,rezende2014stochastic}, the $\beta$-VAE objective \citep{higgins2016beta}, the \(\beta\)-TCVAE objective \citep{chen2018isolating}, and our HFVAE objective.
\todo{Needs nvdm results for \(\beta\)-VAE and \(\beta\)-TCVAE.}

A crucial feature of our experiments is the realization that HFVAE objective serves to induce representations in which correlations in the inference marginal q(z) match correlations in the prior p(z). The two-level decomposition allows us to control the level at which we induce independence. For example, when considering appropriate priors for the MNIST and F-MNIST data, we can take into account the fact that they each contain 10 explicit classes. Likewise for dSprites, containing 3 explicit shape classes. We model these distinct classes using a Concrete distribution \citep{maddison2017concrete,jang2017categorical} of appropriate dimension. We can also assume that these datasets have a multidimensional continuous style variable. HFVAE allows us to induce a stronger independence between the style and the class, while allowing some correlation between the individual style dimensions. We can also use the two-level decomposition to \emph{induce} correlations between variables by \emph{decreasing} the strength of $\term{\textsc{a}}$. As an example, we consider the task of uncovering correlations between topics.


A full list of priors employed is given in Table~\ref{tab:priors}, and the associated model architectures are described in Appendix~\ref{sec:appendix-models}. Note that in connection to Figure \ref{fig:kl-decomp}, subscript `$d$' refers to the variables class or style represented by Concrete and normal distributions respectively, while subscript `$e$' refers to individual dimension \emph{within} the normal variable. In all models, we use a single implementation of the objective based on the Probabilistic Torch \citep{siddharth2017learning} library for deep generative models\footnote{\url{https://github.com/probtorch/probtorch}}.
\begin{table}[b!]
  \vspace*{-2ex}
  \centering
  \begin{tabular}[b]{@{}l@{\hspace{20pt}}c@{\hspace{15pt}}cc@{}}
  \toprule
           & \textbf{VAE}        & \multicolumn{2}{c}{\textbf{HFVAE}}  \\
           & Normal              & Normal              & Concrete      \\
  \midrule
  MNIST    & 10                  & 10                  & 10            \\
  F-MNIST  & 10                  & 10                  & 10            \\
  dSprites & 10                  & 10                  & 3             \\
  CelebA   & 20                  & 20                  & 2             \\
  \bottomrule
  \end{tabular}
  \vspace*{-1ex}
  \caption{Dimensionality of latent variables.}
  \label{tab:priors}
\end{table}

\begin{figure*}[t!]
  \centering
  \begin{tabular}{@{}c@{}c@{}c@{}c@{}c@{}}
    {\small \textsf{Input}}
    & {\small \textsf{$\beta$-VAE}}
    & {\small \textsf{HFVAE}}
    & {\small \textsf{$I(y;z)$}}
    & {\small \textsf{$I(y;z)$}}\\
    & {\tiny \textsf{($\beta=4$)}}
    & {\tiny \textsf{($\beta=12, \gamma=4$)}}
    & {\tiny \textsf{($\beta$-VAE, $\beta=4$)}}
    & {\tiny \textsf{(HFVAE, $\beta=12, \gamma=4$)}}
    \\
    \includegraphics[height=0.14\textheight,valign=t]{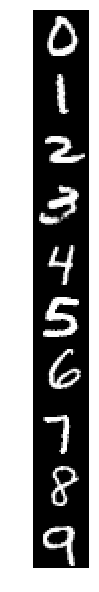}
    & \includegraphics[height=0.14\textheight,valign=t]{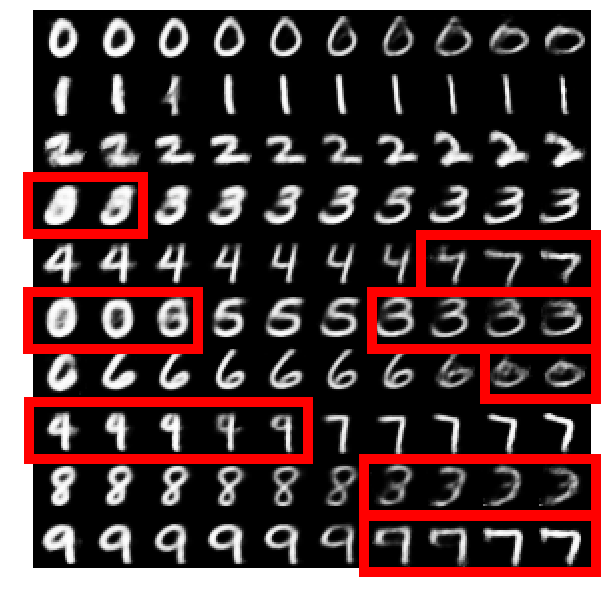}
    & \includegraphics[height=0.14\textheight,valign=t]{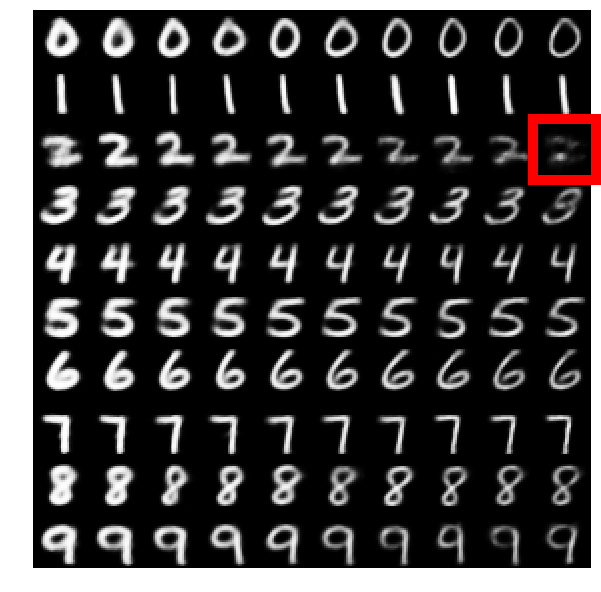}
    & \includegraphics[height=0.14\textheight,valign=t]{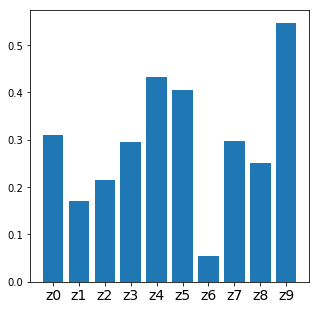}
    & \includegraphics[height=0.14\textheight,valign=t]{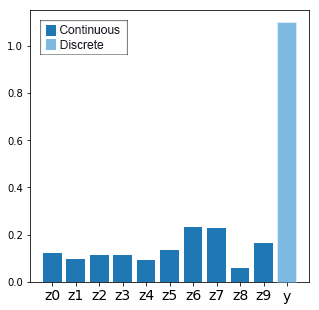}
  \end{tabular}
  \vspace*{-2ex}
  \caption{Left: Manipulation of the thickness variable over the range -3 to 3. The $\beta$-VAE is not able to maintain digit identity as we vary thickness. The HFVAE, which incorporates a discrete variable into the prior is able to maintain the digit identity across the entire range. Right: Mutual information between label $\y$ and individual dimensions of $\z$.}
  \label{fig:mnist-thickness}
\end{figure*}

\begin{figure*}[!t]
  \centering
  \renewcommand{\arraystretch}{0.5}
  \begin{tabular}{@{}c@{}c@{}}
  & {\small \textsf{$I(y;c) - \max_{d}T(y;z_{d})$}} \\
  \includegraphics[height=1.40in,valign=t]{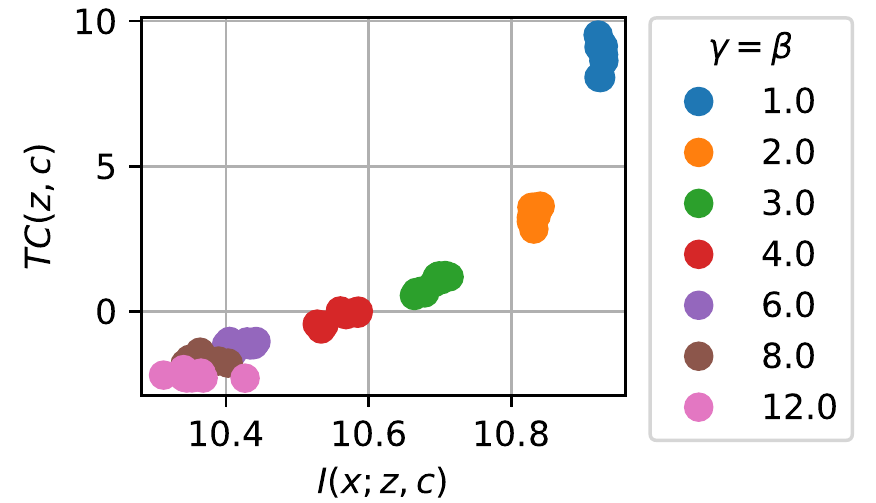}
  & \includegraphics[height=1.40in,valign=t]{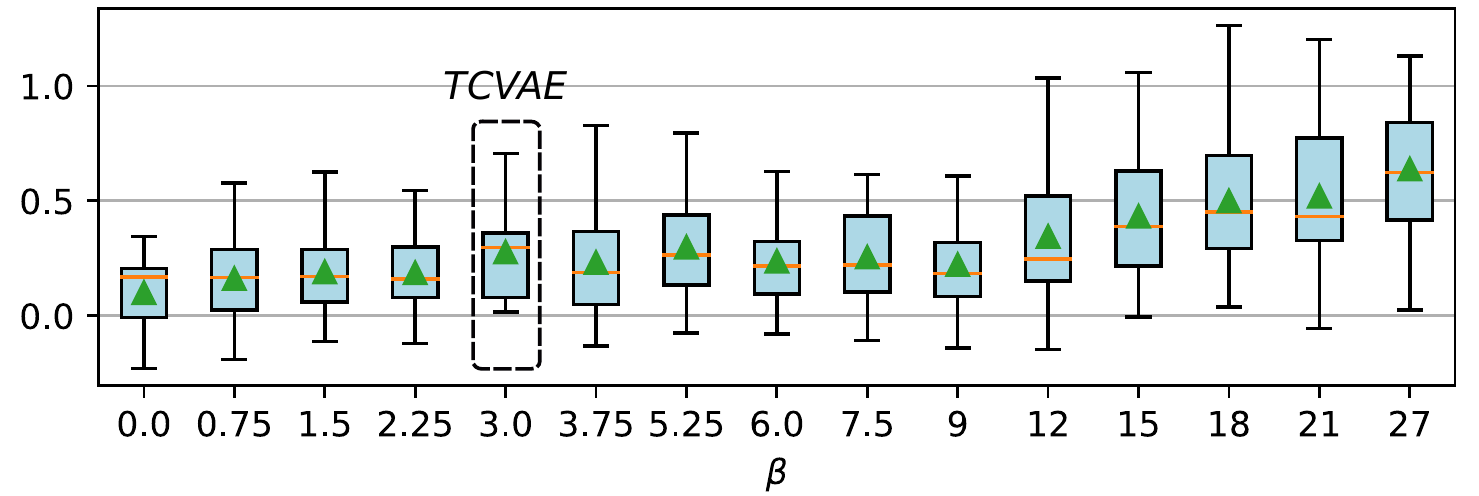}
  \end{tabular}
  \renewcommand{\arraystretch}{1}
  \vspace*{-2ex}
  \caption{On the left, we plot $I(x; \{z,c\})$ vs $TC(\{z, c\})$ for 50 restarts with 7 values $\gamma=\beta$. On the right, we keep $\gamma=3$ fixed and vary $\beta$, plotting mutual information gap (as proposed by \cite{chen2018isolating}) between the Concrete and the largest continuous variable. While there is generally no ``best'' value for $\gamma$, $TC(\{z,c\})$ decreases by 5 nats over range $\gamma=\beta=\{2,3,4,6\}$ while $I(x; \{z,c\})$ decreases by 0.5 nats. Moreover, increasing $\beta$ relative to $\gamma$ also increases the mutual information gap.}
  \label{fig:hyperparam}
  \vspace*{-2ex}
\end{figure*}

\subsection{Qualitative and Quantitative Evaluation}

We begin with a qualitative evaluation of the features that are identified when training with the HFVAE objective. Figure~\ref{fig:mnist-fmnist} shows results for MNIST and F-MNIST datasets. For the MNIST dataset the representation recovers 7 distinct interpretable features---slant, width, height, openness, stroking, thickness, and roundness, while choosing to ignore the remaining dimensions available.
Observing the mutual information~\(I(\x; \z_d)\) between the data~\(\x\) and individual dimensions of the latent space~\(\z_d\) confirms the separation of latent space into useful and ignored subspaces. 

As can be seen from Figure~\ref{fig:mnist-fmnist}(right), the mutual information drops to zero for the unused dimensions. We observe a similar trend for the F-MNIST and CelebA datasets, recovering distinct interpretable factors. For F-MNIST  (Figure~\ref{fig:mnist-fmnist}(mid)), this can be width, length, brightness, etc.\@ For CelebA (Figure~\ref{fig:CelebA}), we uncover interpretable features such as the orientation of the face, variation from smiling to non-smiling, and the presence of sunglasses.

As a quantitative assessment of the quality of learned representations, we evaluate the metrics proposed by \citet{kim2018disentangling} and \citet{eastwood2018framework} on the dSprites dataset with 10 random restarts for each model. For the Eastwood and Williams metric, we used a random forest to regress from features to the (known) ground-truth factors for the data. In Table~\ref{tab:metric-table}, we list these metrics on the dSprites dataset for each of the model types and objectives defined above, noting that the HFVAE performs similar to other approaches.
\begin{table}[b!]
  \centering
  \vspace*{-3ex}
  \begin{tabular}[b]{@{}lcc@{}}
    \toprule
    \textbf{Model}                       & \textbf{Kim}       & \textbf{Eastwood} \\
    \midrule
    VAE                                  & 0.63 $\pm 0.06$    & 0.30 $\pm 0.10$ \\
    $\beta$-VAE ($\beta$ 4.0)            & 0.63 $\pm 0.10$    & 0.41 $\pm 0.11$ \\
    $\beta$-TCVAE ($\beta$ 4.0)          & 0.62 $\pm 0.07$    & 0.29 $\pm 0.10$ \\
    HFVAE ($\beta$ 4.0, $\gamma$ 3.0)    & 0.63 $\pm 0.08$    & 0.39 $\pm 0.16$ \\
    \bottomrule
  \end{tabular}
  \vspace*{-2ex}
  \caption{Disentanglement scores ($\pm$ one standard deviation) for the dSprites dataset using the metrics proposed by \cite{kim2018disentangling} and \cite{eastwood2018framework}.}
  \label{tab:metric-table}
\end{table}

\subsection{Disentangled Representations for Text}

Research on disentangled representations has thus far almost exclusively considered visual (image) data. Exploration of disentangled representations for text is still in its infancy, and generally relies on weak supervision \citep{jain2018learning}.
To assess whether the HFVAE objective can aid interpretability in textual domains, we consider documents in the 20NewsGroups dataset, containing e-mail messages from a number of internet newsgroups; some groups are more closely related  (e.g.~religion vs politics) whereas others are not related at all (e.g.~science vs sports).
We analyze this data using ProdLDA \citep{srivastava2017autoencoding} and neural variational document model (NVDM) \citep{miao2016neural}, both of which are autoencoding neural topic models.
ProdLDA approximates a Dirichlet prior using samples from a Gaussian that are normalized with a softmax function, and then combines this prior with log-linear likelihood model for the words.
NVDM includes a MLP encoder and a log-linear decoder with the assumption of a Gaussian prior.


\begin{figure*}[h!]
  \centering
  \begin{tabular}{@{}c@{}}
    \begin{tabular}{@{}c@{}c@{}}
      \includegraphics[ height=3.8cm]{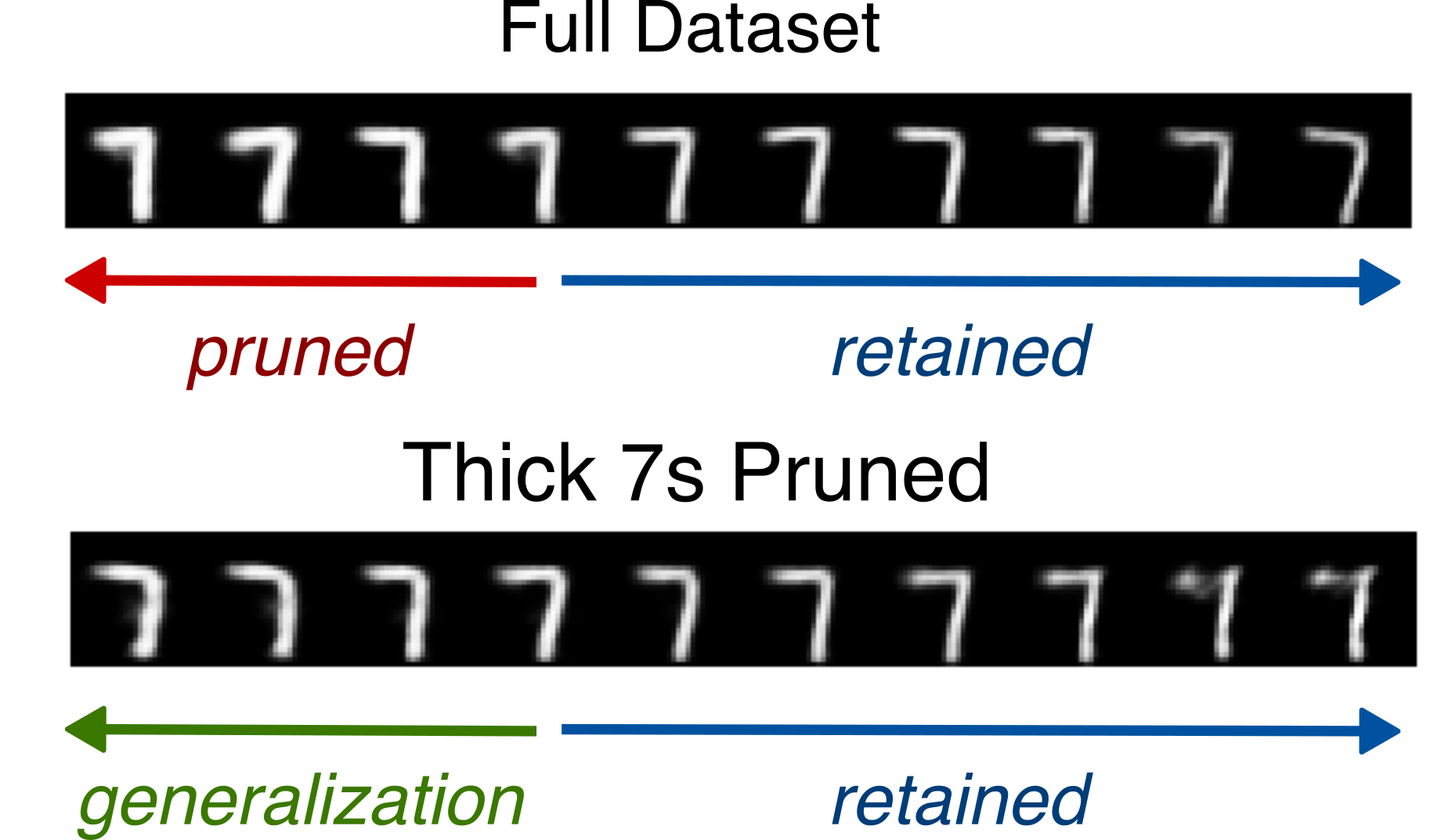}
      & \includegraphics[ height=3.8cm]{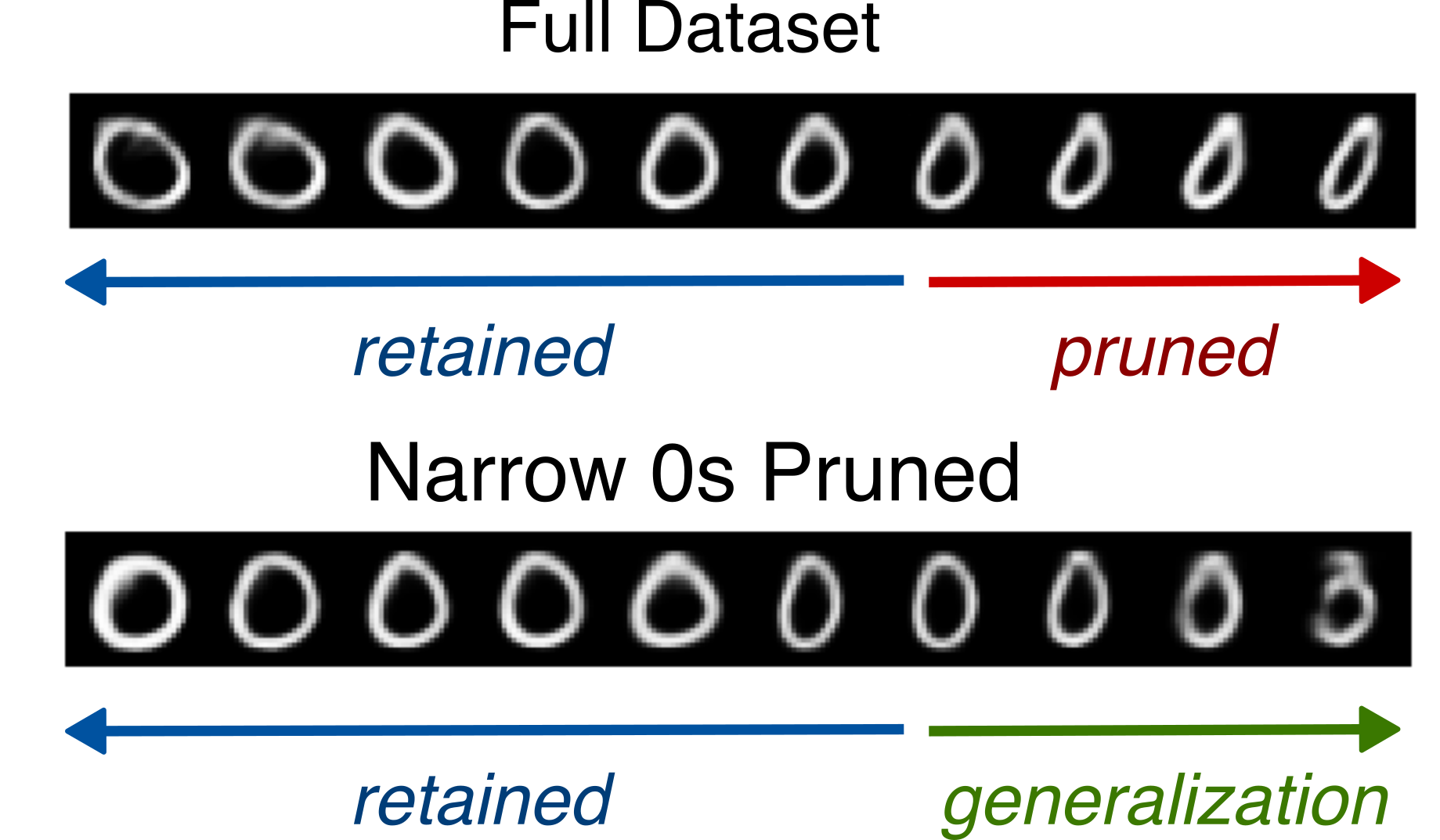}
    \end{tabular}
    \\[5ex]
    \begin{tabular}{@{}c@{\hspace*{-10pt}}c@{\hspace*{-10pt}}c@{\hspace*{-10pt}}c@{\hspace*{-10pt}}}
      \small{\textsf{Pruning}}
      & \small{\textsf{Generalization}}
      & \small{\textsf{Pruning}}
      & \small{\textsf{Generalization}} \\[-0.4ex]
      \includegraphics[width=0.23\linewidth]{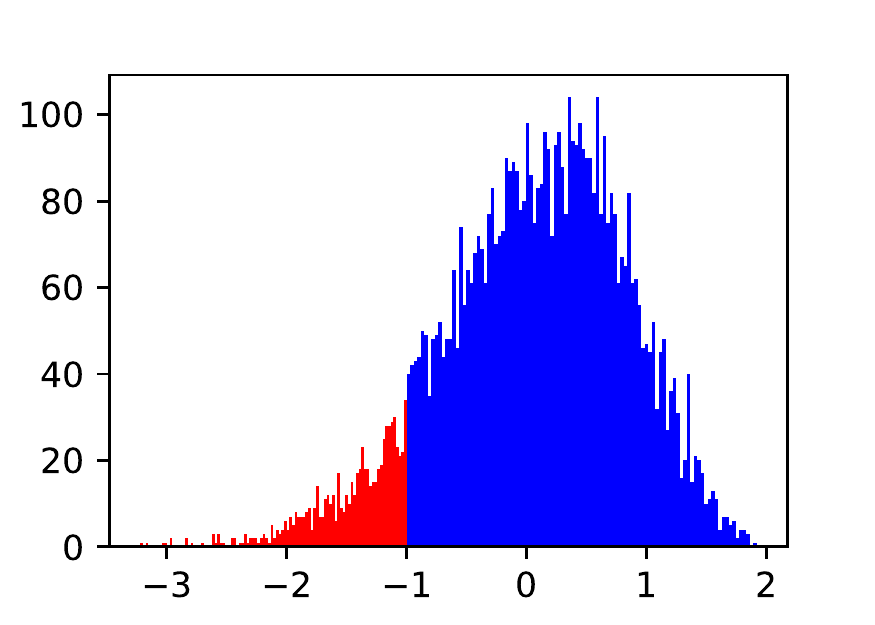}
      & \includegraphics[width=0.23\linewidth]{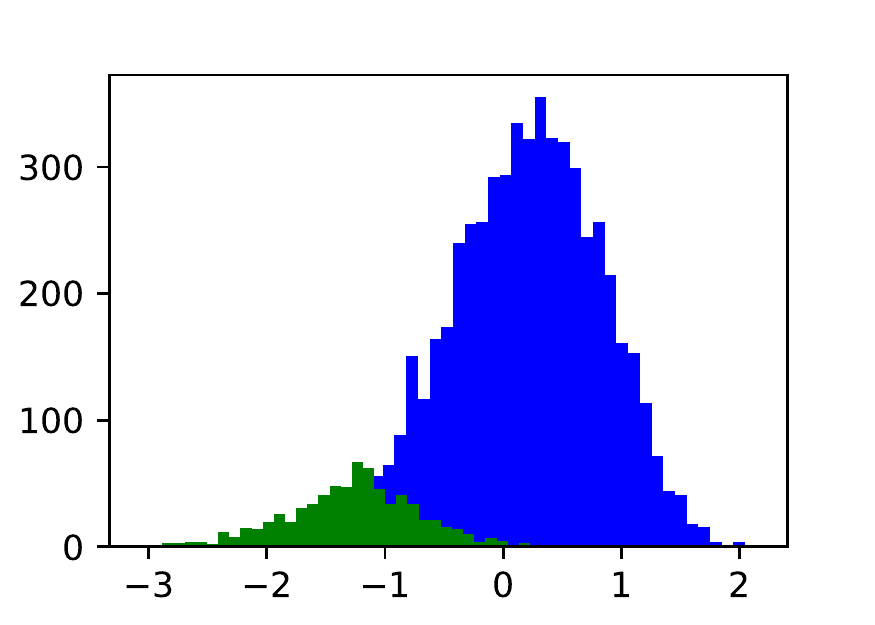}
      & \includegraphics[width=0.23\linewidth]{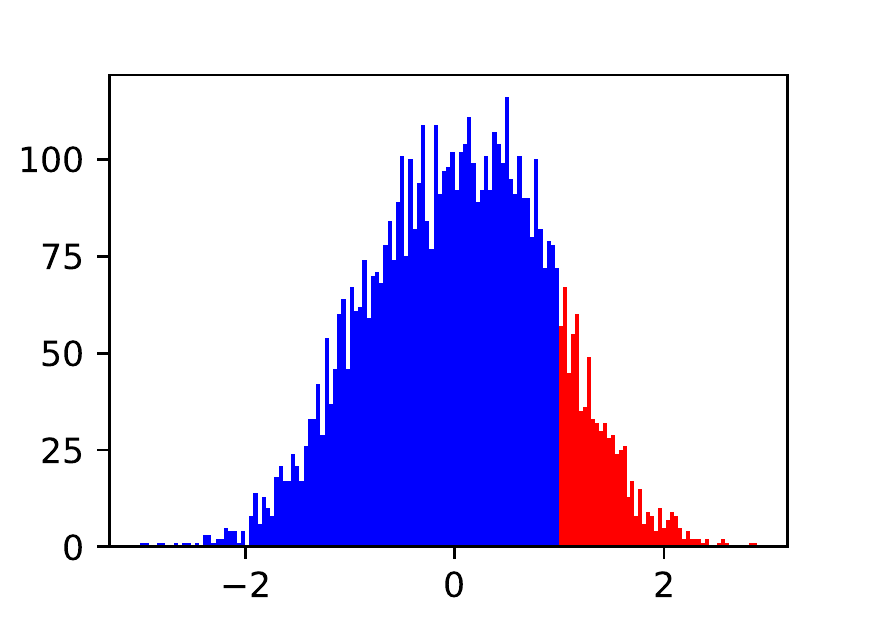}
      & \includegraphics[width=0.23\linewidth]{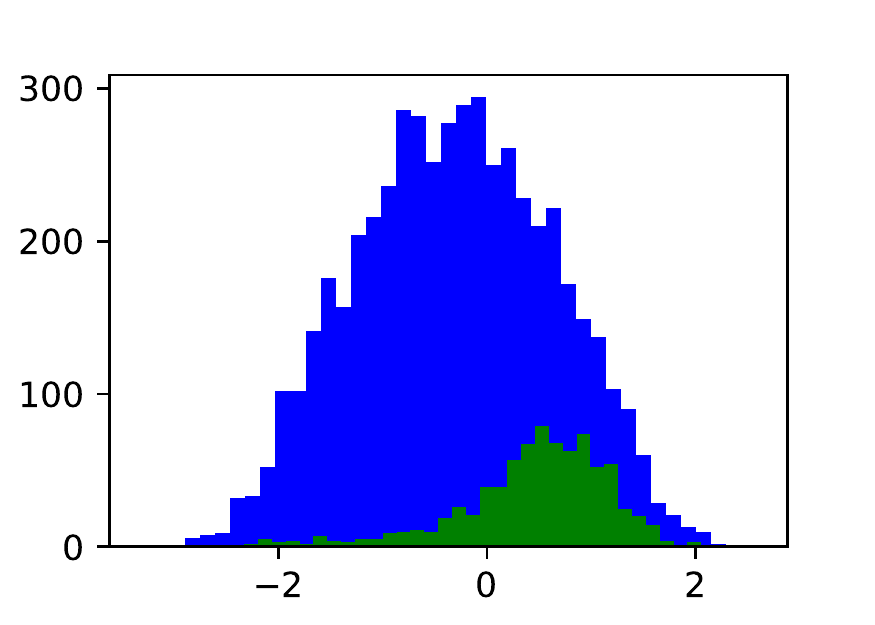}
    \end{tabular}
  \end{tabular}
  \vspace*{-1ex}
  \caption{Generalization to unseen combinations of factors. A HFVAE is trained on the full dataset, and then retrained after a subset of the data is pruned. We then test generalization on the removed portion of the data.}
  \label{fig:mnist-generalization}
  \vspace*{-2ex}
\end{figure*}

To evaluate whether HFVAEs are able to uncover these correlations between topics, we compare a normal ProdLDA model with 50 topics, which we train with a standard VAE objective, to a HFVAE implementation in which we assume two latent variables with 25 dimensions each. We use $\beta = 0.1$ and $\gamma = 4.0$. This means that we \emph{relax} the constraint on the TC at the group level. In other words, the HFVAE model should learn two groups of 25 topics in a manner that \emph{allows} correlations between topics in \emph{different} groups, and \emph{prevents} correlations within a group.

%
Figure~\ref{fig:prod-lda} shows that this approach indeed works as expected. The HFVAE learns correlations between topics that are distinct yet not unrelated, such as religion and politics in the middle east, whereas the VAE does not uncover any significant correlations between topics. As with the MNIST and F-MNIST examples, there is also a significant degree of pruning. The HFVAE learns $11$ topics with a significant mutual information $I(\x; \z_d)$, whereas the VAE has a nonzero  mutual information for all 50 topics.

Additionally, we train an NVDM with a 50-dimensional latent variable using the normal VAE objective,
comparing this baseline to a HFVAE with two 25-dimensional latent variables, trained with $\beta=7$, and $\gamma=4$---\emph{allowing} correlations within a group but \emph{preventing} correlations across groups.
The details of this experiment are in Appendix~\ref{sec:supp-nvdm} and Figure~\ref{fig:nvdm}.
We note that the latent dimensions of the HFVAE achieve a higher degree of disentanglement.

\subsection{Unsupervised Learning of Discrete Labels}

An interesting question relating to the use of discrete latent variables in our framework is how effective these variables are at improving disentanglement. In general, a discrete variable over~\(K\) dimensions expresses a sparsity constraint over those dimensions, as any choice made from that variable is constrained to lie on the vertices of the~\(K\)-dimensional simplex it represents.
This interpretation broadly carries over even in the case of continuous relaxations such as the Concrete distribution that we employ to enable reparameterization for gradient-based methods.

The ability of our framework to disentangle discrete factors of variation \emph{using} the discrete variables depends then, on how separable the underlying classes or identities actually are.
For example, in the case of F-MNIST, a number of classes (i.e., shoes, trousers, dresses, etc.) are visually distinctive enough that they can be captured faithfully by the discrete latent variable.
However, in the case of dSprites, while the shapes (squares, ellipses, and hearts) are conceptually distinct, visually, they can often be difficult to distinguish, with actual differences in the scale of a few pixels.
Here, our approach is less effective at disentangling the relevant factors with the discrete latent.
The MNIST dataset lies in the middle of these two extremes, allowing for clear separation in terms of the digits themselves, but also blurring the lines partially with how similar some of them may appear---a 9 can appear quite similar to a 4, for example, under some small perturbation.

Figure~\ref{fig:mnist-thickness} showcases the ability of the HFVAE to disentangle discrete latent variables from continuous factors of variation in an unsupervised manner.
Here, we have sampled a single data point for each digit and vary the ``thickness''-encoding dimension for each of the $\beta$-VAE and HFVAE models.
Clearly, HFVAE does a better job at disentangling digit vs.\ thickness.
To quantify this ability, as before, we measure the mutual information~\(I(\x; \z_d)\) between the label and each latent dimension, and confirm that the HFVAE learns to encode the information on digits in the discrete variable.
In the case of the $\beta$-VAE, this information is less clearly captured, spread out across the available dimensions.

\subsection{Hyperparameter Analysis}

Here, we analyze the influence of~$\gamma$ and~$\beta$ on the trade-off between total correlation and mutual information. We also show why setting $\gamma \neq \beta$ is necessary for better disentanglement, in comparison to previous objectives \citep{chen2018isolating,kim2018disentangling} where $\gamma = \beta$. We investigate both these facets on the MNIST dataset.
Figure~\ref{fig:hyperparam} (left) shows the results of running our model for 50 restarts, for each of seven different values of \(\gamma = \beta\), indicating positive correlation between mutual information~\(I(x; z,c)\) and total correlation~\(TC(z, c)\). In this figure, the ideal is the lower-right corner, corresponding to high mutual information and low total correlation.
Figure~\ref{fig:hyperparam} (right) shows how the mutual information gap (MIG), defined by \citet{chen2018isolating} to be $I(\y;c) - max_{i}I(\y;\z_{i})$, varies as a function of~\(\beta\), for a fixed~\(\gamma = 3\). We observe that higher values of $\beta$ result in the concrete variable better capturing the label information.

\subsection{Zero-shot Generalization}

A particular feature of disentangled representations is their \emph{utility}, with evidence from human cognition \cite{lake2017building}, suggesting that learning independent factors can aid in generalization to previously unseen combinations of factors.
For example, one can imagine a pink elephant even if one has (sadly) not encountered such an entity previously.

To evaluate if the representations learned using HFVAEs exhibit such properties, we introduce a novel measure of disentanglement quality.
Having first trained a model with the chosen data and objective, here MNIST and the HFVAE, we prune the dataset, removing data containing some particular combinations of factors, say images depicting a thick number 7, or a narrow 0.
We then re-train with the modified dataset, using the pruned data as unseen test data.

Figure~\ref{fig:mnist-generalization} shows the results of this experiment.
As can be seen, the model trained on pruned data is able to successfully reconstruct digits with values for the stroke and character width that were never encountered during training.
The histograms for the feature values show the ability of HFVAE to correctly encode features from previously unseen examples.

\section{Discussion}

Much of the work on learning disentangled representations thus far has focused on cases where the factors of variation are uncorrelated scalar variables. As we begin to apply these techniques to real world datasets, we are likely to encounter correlations between latent variables, particularly when there are causal dependencies between them. This work is a first step towards learning of more structured disentangled representations. By enforcing statistical independence between groups of variables, or relaxing this constraint, we now have the capability to disentangle variables that have higher-dimensional representations. An avenue of future work is to develop datasets that allow us to more rigorously test our ability to characterize correlations between higher-dimensional variables.

\bibliographystyle{plainnat}
{\small\bibliography{refs}}

\newpage
\appendix
\onecolumn
\section{Appendix}

\subsection{Approximating the Inference Marginal}
\label{sec:appendix-qz}

We will here derive a Monte Carlo estimator for the entropy of the marginal $q_\f(\z)$ of the inference model
\begin{align}
	H_\f[\z]
    =
    -
	\E_{q_\f(\z)}
    \left[
      \log
	  q_\f(\z)
    \right].
\end{align}
As with other terms in the objective, we can approximate this expectation by sampling $\z^{b} \sim q_\f(\z)$ using,
\begin{align}
	\x^{b} &\sim q(\x),
	&
	b = 1, \ldots, B,
    \\
    \z^{b}
    &\sim
    q_\f(\z \mid \x^{b}).
\end{align}
We now additionally need to approximate the values,
\begin{align}
	\log q_\f(\z^{b})
    =
    \log
    \left[
    	\frac{1}{N}
    	\sum_{n=1}^N q_\f(\z^{b} \mid \x^n)
    \right].
\end{align}
We will do so by pulling the term for which $\x^n = \x^{b}$ out of the sum
\begin{align*}
	q_\f(\z^{b})
    =
    \frac{1}{N} q_\f(\z^{b} \mid \x^{b})
    +
    \frac{1}{N}
    \sum_{\x^n \not= \x^{b}}
    q_\f(\z^{b} \mid \x^n).
\end{align*}
As also noted by \cite{chen2018isolating}, the intuition behind this decomposition is that $q_\f(\z^{b} \mid \x^{b})$ will in general be much larger than $q_\f(\z^{b} \mid \x^{n})$.

We can approximate the second term using a Monte Carlo estimate from samples $\x^{(b,c)} \sim q(\x \mid \x \not= \x^{b})$,
\begin{align*}
	&
    \frac{1}{N-1}
    \!\!\!\!
    \sum_{\x^n \not= \x^{b}}
    q_\f(\z^{b} \mid \x^n)
	\simeq
    \frac{1}{C}
    \sum_{c=1}^C
    q_\f(\z^{b} \mid \x^{(b,c)}).
\end{align*}
Note here that we have written $1/(N-1)$ instead of $1/N$ in order to ensure that the sum defines an expected value over the distribution $q(\x \mid \x \not= \x^{b})$.

In practice, we can replace the samples $\x^{(b,c)}$ with the samples $b' \not= b$ from the original batch, which yields an estimator over $C=B-1$ samples
\begin{align*}
    \hat{q}(\z^{b})
    =
    \frac{1}{N} q_\f(\z^{b} \mid \x^{b})
    +
    \frac{N-1}{N (B-1)}
    \sum_{b'\not=b}
    q_\f(\z^{b} \mid \x^{b'}).
\end{align*}
Note that this estimator is unbiased, which is to say that
\begin{align}
	\mathbb{E}[\hat{q}(\z^{b})] = q(\z^{b}).
\end{align}
In order to compute the entropy, we now define an estimator $\hat{H}_\f(\z)$, which defines a upper bound on $H_\f(\z)$
\begin{align}
	\hat{H}_\f[\z]
    &\simeq
    -
    \frac{1}{B}
    \sum_{b=1}^B
    \log \hat{q}_\f(\z^{b})
    \ge
    H_\f[\z].
\end{align}
The upper bound relationship follows from Jensen's inequality which states that
\begin{align}
	\mathbb{E}[\log \hat{q}_\f(\z)]
    \le
    \log \mathbb{E}[\hat{q}_\f(\z)]
    =
    \log q_\f(\z).
\end{align}

\subsection{Mutual Information between label \texorpdfstring{$\y$}{y} and representation \texorpdfstring{$\z$}{z}}
\label{sec:supp-mi_y_z}
We quantize each individual dimension $\z_{d}$ into 10 bins based on the CDF of the empirical distribution. In other words, dimension $\z_{d}$ is divided in a way that each bin contains 10\% of the training data. We then compute the mutual information $I(\x; \z_{d})$ as:

\begin{align*}
    I(\z \in \text{bin}_{i}, \y=k) = q(z \in \text{bin}_{i}, \y=k) \left[ \log \frac{q(z \in \text{bin}_{i}, \y=k)}{q(z \in \text{bin}_{i})q(\y=k)} \right]
\end{align*}

For the case where $\z$ is a concrete variable, we use the following formulation:

\begin{align*}
    I(\z=l, \y=k) &= q(\z=l, \y=k) \left[ \log \frac{q(\z=l, \y=k)}{q(\z=l)q(\y=k)} \right] \\
    q(\z=l, \y=k) &= q(\y=k) q(\z=l | \y=k) \\
                &= \frac{N_{k}}{N} q(\z=l | \y=k) \\
    q(\z=l | \y=k) &= \sum_{\x} q(\z=l, \x | \y=k) \\
                         &= \sum_{\x} q(\z=l | \x,  \y=k) q(\x | \y=k) \\
                         &= \frac{1}{N_{k}} \sum_{x} q(\z=l | \x,  \y=k) \\
    q(\z=l) &= \sum_{\x} q(\z=l, \x) \\
           &= \sum_{\x} q(\z=l | \x) q(\x) \\
           &= \frac{1}{N} \sum_{\x} q(\z=l | \x)
\end{align*}

Finally, for the overall mutual information we have:

\begin{align*}
    I(\z, \y) = \sum_{l}\sum_{k} I(\z=l, \y=k)
\end{align*}

\clearpage
\section{Model Architectures}
\label{sec:appendix-models}

We considered 4 datasets:

\textbf{dSprites} \citep{higgins2016beta}: 737,280 binary $64 \times 64$ images of 2D shapes with ground truth factors,

\textbf{MNIST} \citep{lecun2010mnist}: 60000 gray-scale $28 \times 28$ images of handwritten digits,

\textbf{F-MNIST} \citep{xiao2017fashion}: 60000 gray-scale $28 \times 28$ images of clothing items divided in 10 classes,

\textbf{CelebA} \citep{liu2015deep}: 202,599 RGB $64 \times 64 \times 3$ images of celebrity faces. \\

As mentioned in the main text, we used two hidden variables for each of the datasets. One variable is modeled as a normal distribution which representing continuous (denoted as $z_{c}$), and one modeled as a Concrete distribution to detect categories (denoted as $z_{d}$). We used Adam optimizer with learning rate $1e\textit{-3}$ and the default settings.

\vspace{10mm}
\begin{table}[h]
  \label{tab:mnist-architecture}
  \caption{Encoder and Decoder architecture for MNIST and F-MNIST datasets.}
  \centering
  \begin{tabular}{l@{\hspace*{65pt}}l@{\hspace*{65pt}}}
    \toprule
    \textbf{Encoder} & \textbf{Decoder} \\
    \midrule
    Input $28 \times 28$ grayscale image & Input $\z = \textit{Concat}\left[z_{c} \in \mathbb{R}^{10},\ z_{d} \in (0,1)^{10} \right]$ \\
    FC. 400 ReLU & FC. 200 ReLU \\
    FC. $2 \times 200$ ReLU, FC. 10  ($z_{d}$) & FC. 400 ReLU \\
    FC. $2 \times 10$  $(z_{c})$ & FC. $28 \times 28$ Sigmoid\\
    \bottomrule
  \end{tabular}
\end{table}
\vspace{10mm}
\begin{table}[h]
  \label{tab:dSprite-architecture}
  \caption{Encoder and Decoder architecture for dSprites data.}
  \centering
  \begin{tabular}{l@{\hspace*{70pt}}l@{\hspace*{70pt}}}
    \toprule
    \textbf{Encoder} & \textbf{Decoder} \\
    \midrule
    Input $64 \times 64$ binary image & Input $\z = \textit{Concat}\left[z_{c} \in \mathbb{R}^{10}, \ z_{d} \in (0,1)^{3} \right]$ \\
    FC. 1200 ReLU & FC. 400 Tanh \\
    FC. 1200 ReLU & FC. 1200 Tanh \\
    FC. $2 \times 400$ ReLU, FC. 3  ($z_{d}$) & FC. 1200 Tanh \\
    FC. $2 \times 10$  $(z_{c})$ & FC. $64 \times 64$ Sigmoid\\
    \bottomrule
  \end{tabular}
\end{table}
\vspace{10mm}
\begin{table}[h]
  \label{tab:CelebA-architecture}
  \caption{Encoder and Decoder architecture for CelebA data.}
  \centering
  \begin{tabular}{l@{\hspace*{15pt}}l@{\hspace*{15pt}}}
    \toprule
    \textbf{Encoder} & \textbf{Decoder} \\
    \midrule
    Input $64 \times 64$ RGB image & Input $\z = \textit{Concat}\left[z_{c} \in \mathbb{R}^{20}, \ z_{d} \in \{0,1\}^{10} \right]$ \\
    $4 \times 4$ conv, 32 BatchNorm ReLU, stride 2 & FC. 256 ReLU \\
    $4 \times 4$ conv, 32 BatchNorm ReLU, stride 2 & FC. ($4 \times 4 \times 64$) Tanh \\
    $4 \times 4$ conv, 64 BatchNorm ReLU, stride 2 & $4 \times 4$ upconv, 64 BatchNorm ReLU, stride 2 \\
    $4 \times 4$ conv, 64 BatchNorm ReLU, stride 2 & $4 \times 4$ upconv, 32 BatchNorm ReLU, stride 2 \\
    FC. $2 \times 256$ ReLU, 2 FC.  ($z_{d}$) & $4 \times 4$ upconv, 32 BatchNorm ReLU, stride 2 \\
    FC. $2 \times 20$ ReLU & $4 \times 4$ upconv, 3, stride 2 \\
    \bottomrule
  \end{tabular}
\end{table}

\clearpage

\section{Latent Traversals}
\begin{figure}[h!]
 \centering
 \includegraphics[width=0.55\linewidth]{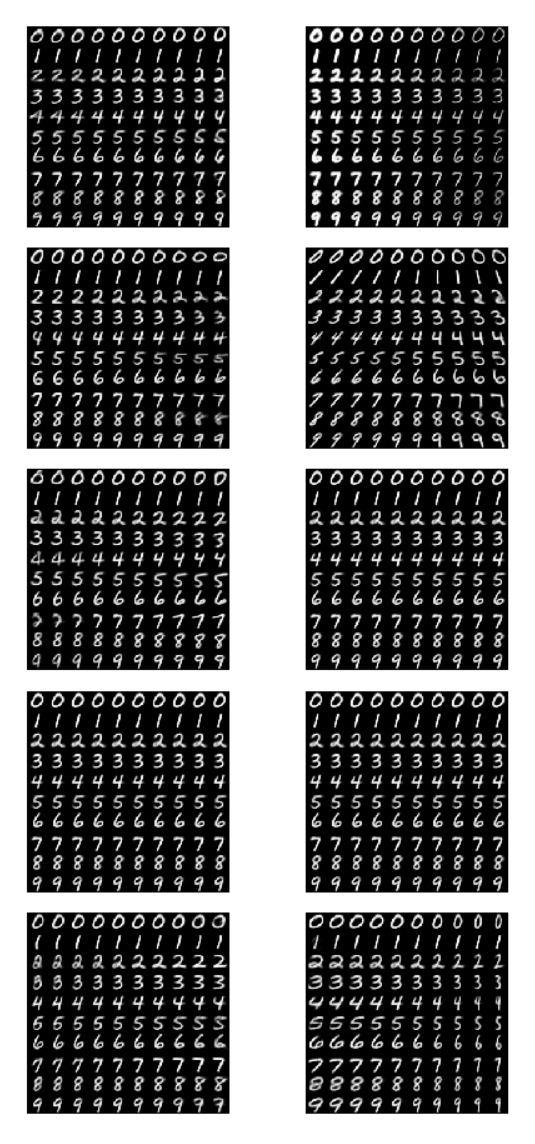}
 \label{fig:mnist-full-result}
 \caption{Qualitative results for disentanglement in MNIST dataset. In each case, one particular $z_{d}$ is varying from -3 to 3 while the others are fixed at 0. For this particular set of traversals, we used 10\% supervision in order to extract the digit more reliably, therefore visualizing all `style' features present in MNIST.}
\end{figure}

\begin{figure}
 \centering
 \includegraphics[width=0.5\linewidth]{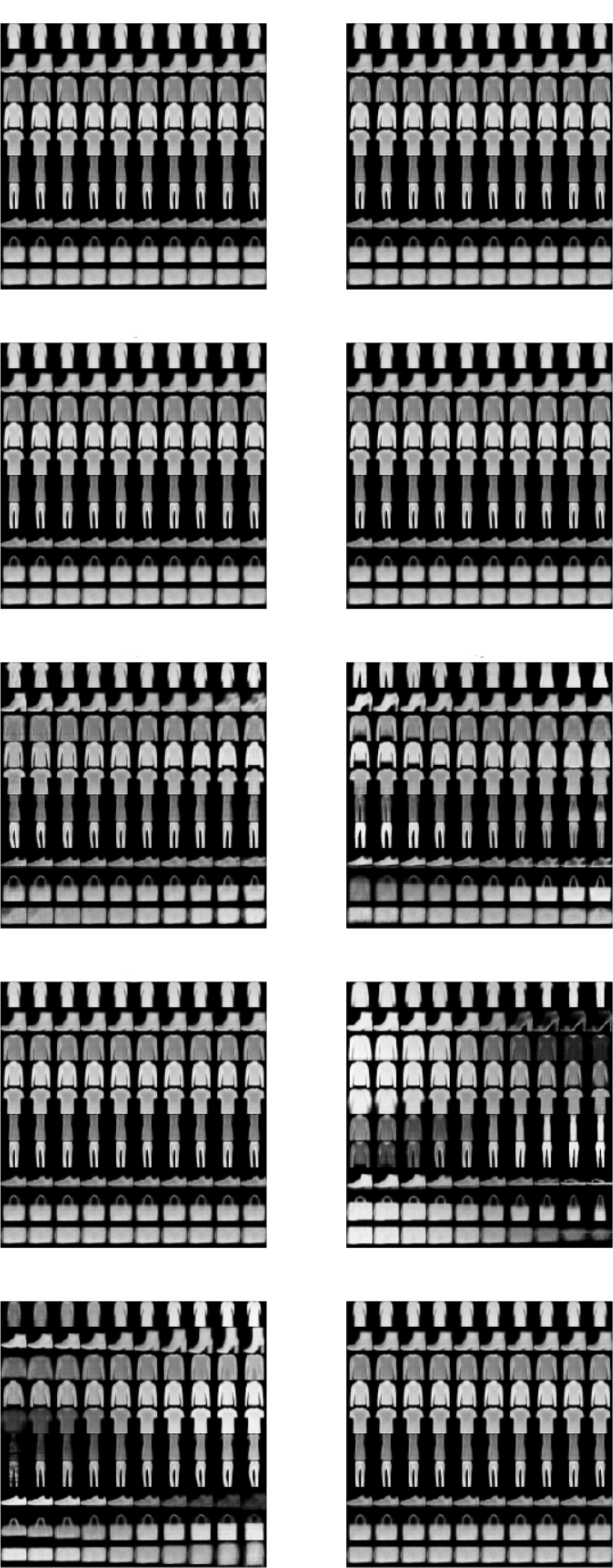}
 \label{fig:app-mnist-full-result}
 \caption{Qualitative results for disentanglement in F-MNIST dataset. In each case, one particular $z_{d}$ is varying from -3 to 3 while the others are fixed at 0}
\end{figure}

\clearpage
\section{Disentangled Representation for Text}

\subsection{Model Architectures}

We consider the following dataset:

\textbf{20NewsGroups} : 11314 newsgroup documents which are partitioned in 20 categories. We used bag-of-words representation where vocabulary size is 2000, after removing stopwords using Mallet stopwords list.\\

With HFVAE objective, we used two hidden variables (denoted as $z_{c1}$ and $z_{c2}$) with 25 dimensions each. In ProdLDA, we used Adam optimizer with $\beta_1 = 0.99$, $\beta_2 = 0.999$, and learning rate $1e\textit{-3}$; In NVDM, we used Adam optimizer with learning rate $5e\textit{-5}$ and default settings.

\vspace{10mm}
\begin{table}[h!]
  \centering
  \begin{tabular}{l@{\hspace*{65pt}}l@{\hspace*{65pt}}}
    \toprule
    \textbf{Encoder} & \textbf{Decoder} \\
    \midrule
    Input $1 \times 2000$ document & Input $z_{c1} \in \mathbb{R}^{25}$, $z_{c2} \in \mathbb{R}^{25}$ \\
    FC. 100 Softplus & Softmax Dropout \\
    FC. 100 Softplus Dropout & FC. 2000 BatchNorm Softmax \\
    FC. $2 \times 25$ BatchNorm $(z_{c1})$ \\
    FC. $2 \times 25$ BatchNorm $(z_{c2})$ \\
    \bottomrule
  \end{tabular}
  \caption{Encoder and Decoder architecture in ProdLDA.}
  \label{tab:prodlda-architecture}
\end{table}

\vspace{10mm}
\begin{table}[h!]
  \centering
  \begin{tabular}{l@{\hspace*{65pt}}l@{\hspace*{65pt}}}
    \toprule
    \textbf{Encoder} & \textbf{Decoder} \\
    \midrule
    Input $1 \times 2000$ document & Input $z_{c1} \in \mathbb{R}^{25}$, $z_{c2} \in \mathbb{R}^{25}$ \\
    FC. 500 ReLU & FC. 2000 Softmax\\
    FC. $2 \times 25$  $(z_{c1})$ \\
    FC. $2 \times 25$  $(z_{c2})$ \\
    \bottomrule
  \end{tabular}
  \caption{Encoder and Decoder architecture in NVDM.}
  \label{tab:nvdm-architecture}
\end{table}

\subsection{Neural variational document model}
\label{sec:supp-nvdm}

\begin{figure*}[!t]
    \centering
    \includegraphics[width=\textwidth]{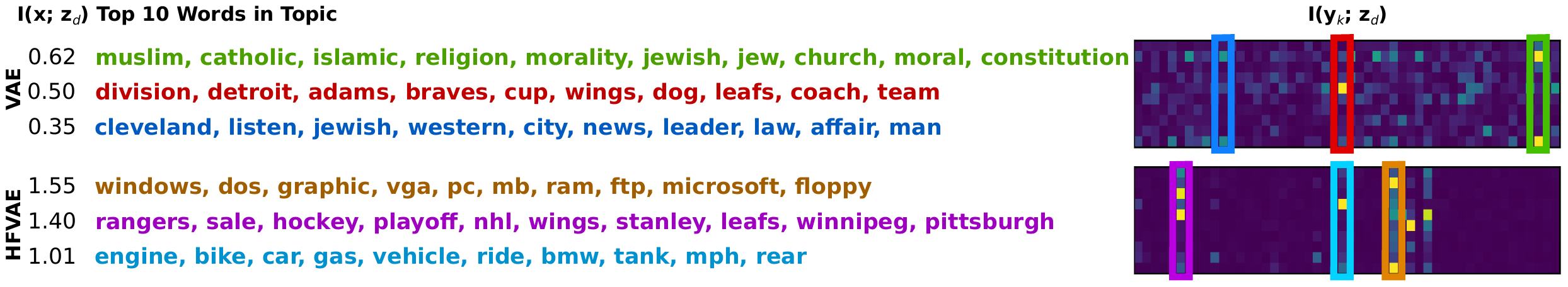}
    \caption{Learned topics in 20NewsGroups dataset using HFVAE objective and VAE objective. The middle column shows frequent words for the 3 most informative dimensions of the latent space. The left column lists their corresponding mutual information with $\x$. The right column shows the mutual information between the latent code and binary indicator variable for the document category.}
    \label{fig:nvdm}
    \vspace*{-2ex}
\end{figure*}

We train a standard NVDM with a 50-dimensional latent variable using the normal VAE objective. We compare this baseline to a HFVAE with two 25-dimensional latent variables, trained with $\beta=7$, and $\gamma=4$---\emph{allowing} correlations within a group but \emph{preventing} correlations across groups.

Figure~\ref{fig:nvdm} (right) shows the mutual information between the latent code and binary indicator variables for the document category. We see that the latent dimensions of the HFVAE (columns) achieve a higher degree of disentanglement as is evident from the fact that indicator labels (shown as rows) correlate generally with only one latent feature (shown in columns). Note that a single feature can capture two distinct topics in this model (of which only one is shown), which correspond to negative and positive weights in the likelihood model.

\subsection{Binary Indicator Variables for Document Category}

In 20NewsGroups dataset, we derived 10 binary variables where each indicates whether a document belongs to this specific topic. Since some of the newsgroups are closely related (e.g.~religion vs politics) while others are not related at all (e.g.~science vs sports), we regarded highly related categories as one single topic. Then we computed the mutual information between each binary indicator variable $b_l$ and individual dimension $z_d$ (see Appendix~\ref{sec:supp-mi_y_z}), which is shown in Figure~\ref{fig:nvdm}.

\begin{table}[h!]
  \label{tab:20newsgroups-binary_variable}
  \caption{Topics after grouping highly related categories.}
  \centering
  \begin{tabular}{l@{\hspace*{15pt}}l@{\hspace*{15pt}}}
    \toprule
    \textbf{Grouped Topics} & \textbf{Original Categories} \\
    \midrule
    Atheism & alt.atheism \\
    \hline
    Computer & comp.graphics \\
             & comp.os.ms-windows.misc \\
             & comp.sys.ibm.pc.hardware \\
             & comp.sys.mac.hardware \\
             & comp.windows.x \\
    \hline
    Forsale & misc.forsale \\
    \hline
    Autos & rec.autos \\
          & rec.motorcycles \\
    \hline
    Sports & rec.sport.baseball \\
           & rec.sport.hockey \\
    \hline
    Encryption & sci.crypt \\
    \hline
    Electronics & sci.electronics \\
    \hline
    Medical & sci.med \\
    \hline
    Space &  sci.space\\
    \hline
    Politics and Religion & talk.politics.misc \\
                           & talk.politics.guns \\
                           & talk.politics.mideast \\
                           & talk.religion.misc \\
                           & soc.religion.christian \\
    \bottomrule
  \end{tabular}
\end{table}

\end{document}

%% file: macros.tex
\newcommand{\argmax}{\operatornamewithlimits{argmax}}
\newcommand{\argmin}{\operatornamewithlimits{argmin}}

\let\avec\vec
\renewcommand{\vec}[1]{\ensuremath{\boldsymbol{#1}}}
\renewcommand{\v}[1]{\vec{#1}}

\newcommand{\x}{\ensuremath{\v{x}}}
\newcommand{\y}{\ensuremath{\v{y}}}
\newcommand{\z}{\ensuremath{\v{z}}}
\newcommand{\h}{\ensuremath{\v{\eta}}}
\renewcommand{\u}{\ensuremath{u}}
\newcommand{\q}{\ensuremath{\theta}}
\newcommand{\f}{\ensuremath{\phi}}
\renewcommand{\l}{\ensuremath{\lambda}}
\renewcommand{\t}{\ensuremath{\tau}}

\renewcommand{\L}{\ensuremath{\mathcal{L}}}
\newcommand{\KL}[2]{\ensuremath{\text{KL}\left({#1} \:\middle\vert\middle\vert\: {#2}\right)}}
\newcommand{\E}{\ensuremath{\mathbb{E}}}
\newcommand{\N}{\ensuremath{\mathcal{N}}}
\newcommand{\C}{\ensuremath{\mathtt{Concrete}}}

%% file: aistats_2019.bbl
\begin{thebibliography}{46}
\providecommand{\natexlab}[1]{#1}
\providecommand{\url}[1]{\texttt{#1}}
\expandafter\ifx\csname urlstyle\endcsname\relax
  \providecommand{\doi}[1]{doi: #1}\else
  \providecommand{\doi}{doi: \begingroup \urlstyle{rm}\Url}\fi

\bibitem[Achille and Soatto(2018)]{achille2018information}
A.~Achille and S.~Soatto.
\newblock Information {{Dropout}}: {{Learning Optimal Representations Through
  Noisy Computation}}.
\newblock \emph{IEEE Transactions on Pattern Analysis and Machine
  Intelligence}, PP\penalty0 (99):\penalty0 1--1, 2018.

\bibitem[Alemi et~al.(2018)Alemi, Poole, Fischer, Dillon, Saurous, and
  Murphy]{alemi2018fixing}
Alexander Alemi, Ben Poole, Ian Fischer, Joshua Dillon, Rif~A Saurous, and
  Kevin Murphy.
\newblock Fixing a broken elbo.
\newblock In \emph{International Conference on Machine Learning}, pages
  159--168, 2018.

\bibitem[Alemi et~al.(2016)Alemi, Fischer, Dillon, and Murphy]{alemi2016deep}
Alexander~A. Alemi, Ian Fischer, Joshua~V. Dillon, and Kevin Murphy.
\newblock Deep {{Variational Information Bottleneck}}.
\newblock \emph{arXiv:1612.00410 [cs, math]}, December 2016.

\bibitem[Bouchacourt et~al.(2017)Bouchacourt, Tomioka, and
  Nowozin]{bouchacourt2017multi}
Diane Bouchacourt, Ryota Tomioka, and Sebastian Nowozin.
\newblock Multi-level variational autoencoder: Learning disentangled
  representations from grouped observations.
\newblock \emph{arXiv preprint arXiv:1705.08841}, 2017.

\bibitem[Bowman et~al.(2016)Bowman, Vilnis, Vinyals, Dai, Jozefowicz, and
  Bengio]{bowman2016generating}
Samuel~R Bowman, Luke Vilnis, Oriol Vinyals, Andrew~M Dai, Rafal Jozefowicz,
  and Samy Bengio.
\newblock Generating sentences from a continuous space.
\newblock \emph{CoNLL 2016}, page~10, 2016.

\bibitem[Burgess et~al.(2018)Burgess, Higgins, Pal, Matthey, Watters,
  Desjardins, and Lerchner]{burgess2018understanding}
Christopher~P Burgess, Irina Higgins, Arka Pal, Loic Matthey, Nick Watters,
  Guillaume Desjardins, and Alexander Lerchner.
\newblock Understanding disentangling in $\beta$-vae.
\newblock \emph{arXiv preprint arXiv:1804.03599}, 2018.

\bibitem[Chen et~al.(2018)Chen, Li, Grosse, and Duvenaud]{chen2018isolating}
Tian~Qi Chen, Xuechen Li, Roger Grosse, and David Duvenaud.
\newblock Isolating sources of disentanglement in variational autoencoders.
\newblock \emph{arXiv preprint arXiv:1802.04942}, 2018.

\bibitem[Chen et~al.(2016{\natexlab{a}})Chen, Duan, Houthooft, Schulman,
  Sutskever, and Abbeel]{chen2016infogan}
Xi~Chen, Yan Duan, Rein Houthooft, John Schulman, Ilya Sutskever, and Pieter
  Abbeel.
\newblock Infogan: {{Interpretable}} representation learning by information
  maximizing generative adversarial nets.
\newblock In \emph{Advances in {{Neural Information Processing Systems}}},
  pages 2172--2180, 2016{\natexlab{a}}.

\bibitem[Chen et~al.(2016{\natexlab{b}})Chen, Kingma, Salimans, Duan, Dhariwal,
  Schulman, Sutskever, and Abbeel]{chen2016variational}
Xi~Chen, Diederik~P. Kingma, Tim Salimans, Yan Duan, Prafulla Dhariwal, John
  Schulman, Ilya Sutskever, and Pieter Abbeel.
\newblock Variational lossy autoencoder.
\newblock \emph{arXiv preprint arXiv:1611.02731}, 2016{\natexlab{b}}.

\bibitem[Donahue et~al.(2016)Donahue, Kr{\"a}henb{\"u}hl, and
  Darrell]{donahue2016bigan}
Jeff Donahue, Philipp Kr{\"a}henb{\"u}hl, and Trevor Darrell.
\newblock Adversarial feature learning.
\newblock \emph{arXiv preprint arXiv:1605.09782}, 2016.

\bibitem[Dumoulin et~al.(2016)Dumoulin, Belghazi, Poole, Mastropietro, Lamb,
  Arjovsky, and Courville]{dumoulin2016ali}
Vincent Dumoulin, Ishmael Belghazi, Ben Poole, Olivier Mastropietro, Alex Lamb,
  Martin Arjovsky, and Aaron Courville.
\newblock Adversarially learned inference.
\newblock \emph{arXiv preprint arXiv:1606.00704}, 2016.

\bibitem[Dupont(2018)]{dupont2018joint}
Emilien Dupont.
\newblock Joint-vae: Learning disentangled joint continuous and discrete
  representations.
\newblock \emph{arXiv preprint arXiv:1804.00104}, 2018.

\bibitem[Eastwood and Williams(2018)]{eastwood2018framework}
Cian Eastwood and Christopher K.~I. Williams.
\newblock A {{Framework}} for the {{Quantitative Evaluation}} of {{Disentangled
  Representations}}.
\newblock In \emph{International Conference on Learning Representations},
  February 2018.

\bibitem[Engel et~al.(2017)Engel, Hoffman, and Roberts]{engel2017latent}
Jesse Engel, Matthew Hoffman, and Adam Roberts.
\newblock Latent {{Constraints}}: {{Learning}} to {{Generate Conditionally}}
  from {{Unconditional Generative Models}}.
\newblock \emph{arXiv:1711.05772 [cs, stat]}, November 2017.

\bibitem[Gao et~al.(2018)Gao, Brekelmans, Steeg, and Galstyan]{gao2018auto}
Shuyang Gao, Rob Brekelmans, Greg~Ver Steeg, and Aram Galstyan.
\newblock Auto-encoding total correlation explanation.
\newblock \emph{arXiv preprint arXiv:1802.05822}, 2018.

\bibitem[Gatys et~al.(2015)Gatys, Ecker, and Bethge]{gatys2015neural}
Leon~A Gatys, Alexander~S Ecker, and Matthias Bethge.
\newblock A neural algorithm of artistic style.
\newblock \emph{arXiv preprint arXiv:1508.06576}, 2015.

\bibitem[G{\'o}mez-Bombarelli et~al.(2018)G{\'o}mez-Bombarelli, Wei, Duvenaud,
  Hern{\'a}ndez-Lobato, S{\'a}nchez-Lengeling, Sheberla, Aguilera-Iparraguirre,
  Hirzel, Adams, and Aspuru-Guzik]{gomez2018automatic}
Rafael G{\'o}mez-Bombarelli, Jennifer~N Wei, David Duvenaud, Jos{\'e}~Miguel
  Hern{\'a}ndez-Lobato, Benjam{\'\i}n S{\'a}nchez-Lengeling, Dennis Sheberla,
  Jorge Aguilera-Iparraguirre, Timothy~D Hirzel, Ryan~P Adams, and Al{\'a}n
  Aspuru-Guzik.
\newblock Automatic chemical design using a data-driven continuous
  representation of molecules.
\newblock \emph{ACS Central Science}, 2018.

\bibitem[Gulrajani et~al.(2017)Gulrajani, Kumar, Ahmed, Taiga, Visin, Vazquez,
  and Courville]{gulrajani2016pixelvae}
Ishaan Gulrajani, Kundan Kumar, Faruk Ahmed, Adrien~Ali Taiga, Francesco Visin,
  David Vazquez, and Aaron Courville.
\newblock {PixelVAE}: A latent variable model for natural images.
\newblock In \emph{International Conference on Machine Learning}, 2017.

\bibitem[Higgins et~al.(2016)Higgins, Matthey, Pal, Burgess, Glorot, Botvinick,
  Mohamed, and Lerchner]{higgins2016beta}
Irina Higgins, Loic Matthey, Arka Pal, Christopher Burgess, Xavier Glorot,
  Matthew Botvinick, Shakir Mohamed, and Alexander Lerchner.
\newblock beta-{VAE}: Learning basic visual concepts with a constrained
  variational framework.
\newblock In \emph{International Conference on Learning Representations}, 2016.

\bibitem[Hoffman and Johnson(2016)]{hoffman2016elbo}
Matthew~D. Hoffman and Matthew~J. Johnson.
\newblock Elbo surgery: Yet another way to carve up the variational evidence
  lower bound.
\newblock In \emph{Workshop in {{Advances}} in {{Approximate Bayesian
  Inference}}, {{NIPS}}}, 2016.

\bibitem[Jain et~al.(2018)Jain, Banner, {van de Meent}, Marshall, and
  Wallace]{jain2018learning}
Sarthak Jain, Edward Banner, Jan-Willem {van de Meent}, Iain~J. Marshall, and
  Byron~C. Wallace.
\newblock Learning {{Disentangled Representations}} of {{Texts}} with
  {{Application}} to {{Biomedical Abstracts}}.
\newblock \emph{arXiv:1804.07212 [cs]}, April 2018.

\bibitem[Jang et~al.(2017)Jang, Gu, and Poole]{jang2017categorical}
Eric Jang, Shixiang Gu, and Ben Poole.
\newblock Categorical reparameterization with {G}umbel-softmax.
\newblock In \emph{International Conference on Learning Representations}, 2017.

\bibitem[Karaletsos et~al.(2015)Karaletsos, Belongie, and
  R{\"a}tsch]{karaletsos2015bayesian}
Theofanis Karaletsos, Serge Belongie, and Gunnar R{\"a}tsch.
\newblock Bayesian representation learning with oracle constraints.
\newblock \emph{arXiv preprint arXiv:1506.05011}, 2015.

\bibitem[Kim and Mnih(2018)]{kim2018disentangling}
Hyunjik Kim and Andriy Mnih.
\newblock Disentangling by factorising.
\newblock \emph{arXiv preprint arXiv:1802.05983}, 2018.

\bibitem[Kingma and Welling(2013)]{kingma2013auto-encoding}
Diederik~P. Kingma and Max Welling.
\newblock Auto-encoding variational bayes.
\newblock In \emph{International Conference on Learning Representations}, 2013.

\bibitem[Kingma et~al.(2014)Kingma, Mohamed, Rezende, and
  Welling]{kingma2014semi}
Diederik~P Kingma, Shakir Mohamed, Danilo~Jimenez Rezende, and Max Welling.
\newblock Semi-supervised learning with deep generative models.
\newblock In \emph{Advances in Neural Information Processing Systems}, pages
  3581--3589, 2014.

\bibitem[Kulkarni et~al.(2015)Kulkarni, Whitney, Kohli, and
  Tenenbaum]{kulkarni2015deep}
Tejas~D Kulkarni, William~F Whitney, Pushmeet Kohli, and Josh Tenenbaum.
\newblock Deep convolutional inverse graphics network.
\newblock In \emph{Advances in Neural Information Processing Systems}, pages
  2539--2547, 2015.

\bibitem[Kumar et~al.(2017)Kumar, Sattigeri, and
  Balakrishnan]{kumar2017variational}
Abhishek Kumar, Prasanna Sattigeri, and Avinash Balakrishnan.
\newblock Variational inference of disentangled latent concepts from unlabeled
  observations.
\newblock \emph{arXiv preprint arXiv:1711.00848}, 2017.

\bibitem[Kusner et~al.(2017)Kusner, Paige, and
  Hern{\'a}ndez-Lobato]{kusner2017grammar}
Matt~J Kusner, Brooks Paige, and Jos{\'e}~Miguel Hern{\'a}ndez-Lobato.
\newblock Grammar variational autoencoder.
\newblock In \emph{International Conference on Machine Learning}, 2017.

\bibitem[Lake et~al.(2017)Lake, Ullman, Tenenbaum, and
  Gershman]{lake2017building}
Brenden~M Lake, Tomer~D Ullman, Joshua~B Tenenbaum, and Samuel~J Gershman.
\newblock Building machines that learn and think like people.
\newblock \emph{Behavioral and Brain Sciences}, 40, 2017.

\bibitem[Lang(2007)]{20newsgroups}
Ken Lang.
\newblock {20 newsgroups data set}, 2007.
\newblock URL \url{http://www.ai.mit.edu/people/jrennie/20Newsgroups/}.
\newblock [Online; accessed 18-May-2018].

\bibitem[LeCun et~al.(2010)LeCun, Cortes, and Burges]{lecun2010mnist}
Yann LeCun, Corinna Cortes, and CJ~Burges.
\newblock Mnist handwritten digit database.
\newblock \emph{AT\&T Labs [Online]. Available: http://yann. lecun.
  com/exdb/mnist}, 2, 2010.

\bibitem[Liang et~al.(2018)Liang, Krishnan, Hoffman, and
  Jebara]{liang2018variational}
Dawen Liang, Rahul~G. Krishnan, Matthew~D. Hoffman, and Tony Jebara.
\newblock Variational {{Autoencoders}} for {{Collaborative Filtering}}.
\newblock \emph{arXiv:1802.05814 [cs, stat]}, February 2018.

\bibitem[Liu et~al.(2015)Liu, Luo, Wang, and Tang]{liu2015deep}
Ziwei Liu, Ping Luo, Xiaogang Wang, and Xiaoou Tang.
\newblock Deep learning face attributes in the wild.
\newblock In \emph{Proceedings of the {{IEEE International Conference}} on
  {{Computer Vision}}}, pages 3730--3738, 2015.

\bibitem[Maddison et~al.(2017)Maddison, Mnih, and Teh]{maddison2017concrete}
Chris~J Maddison, Andriy Mnih, and Yee~Whye Teh.
\newblock The concrete distribution: A continuous relaxation of discrete random
  variables.
\newblock In \emph{International Conference on Learning Representations}, 2017.

\bibitem[Makhzani et~al.(2015)Makhzani, Shlens, Jaitly, Goodfellow, and
  Frey]{makhzani2015adversarial}
Alireza Makhzani, Jonathon Shlens, Navdeep Jaitly, Ian Goodfellow, and Brendan
  Frey.
\newblock Adversarial autoencoders.
\newblock \emph{arXiv preprint arXiv:1511.05644}, 2015.

\bibitem[Miao et~al.(2016)Miao, Yu, and Blunsom]{miao2016neural}
Yishu Miao, Lei Yu, and Phil Blunsom.
\newblock Neural variational inference for text processing.
\newblock In \emph{International Conference on Machine Learning}, pages
  1727--1736, 2016.

\bibitem[Oord et~al.(2016)Oord, Dieleman, Zen, Simonyan, Vinyals, Graves,
  Kalchbrenner, Senior, and Kavukcuoglu]{oord2016wavenet}
Aaron van~den Oord, Sander Dieleman, Heiga Zen, Karen Simonyan, Oriol Vinyals,
  Alex Graves, Nal Kalchbrenner, Andrew Senior, and Koray Kavukcuoglu.
\newblock Wavenet: A generative model for raw audio.
\newblock \emph{arXiv preprint arXiv:1609.03499}, 2016.

\bibitem[Rezende et~al.(2014)Rezende, Mohamed, and
  Wierstra]{rezende2014stochastic}
Danilo~Jimenez Rezende, Shakir Mohamed, and Daan Wierstra.
\newblock Stochastic backpropagation and approximate inference in deep
  generative models.
\newblock In \emph{Proceedings of The 31st International Conference on Machine
  Learning}, pages 1278--1286, 2014.

\bibitem[Siddharth et~al.(2017)Siddharth, Paige, Van~de Meent, Desmaison, Wood,
  Goodman, Kohli, and Torr]{siddharth2017learning}
N~Siddharth, Brooks Paige, Jan-Willem Van~de Meent, Alban Desmaison, Frank
  Wood, Noah~D Goodman, Pushmeet Kohli, and Philip~HS Torr.
\newblock Learning disentangled representations with semi-supervised deep
  generative models.
\newblock In \emph{Advances in Neural Information Processing Systems}, 2017.

\bibitem[Srivastava and Sutton(2017)]{srivastava2017autoencoding}
Akash Srivastava and Charles Sutton.
\newblock Autoencoding variational inference for topic models.
\newblock \emph{arXiv preprint arXiv:1703.01488}, 2017.

\bibitem[Theis et~al.(2017)Theis, Shi, Cunningham, and
  Husz{\'a}r]{theis2017lossy}
Lucas Theis, Wenzhe Shi, Andrew Cunningham, and Ferenc Husz{\'a}r.
\newblock Lossy image compression with compressive autoencoders.
\newblock \emph{arXiv preprint arXiv:1703.00395}, 2017.

\bibitem[Veit et~al.(2016)Veit, Belongie, and
  Karaletsos]{veit2016disentangling}
Andreas Veit, Serge Belongie, and Theofanis Karaletsos.
\newblock Disentangling {{Nonlinear Perceptual Embeddings With Multi}}-{{Query
  Triplet Networks}}.
\newblock \emph{arXiv preprint arXiv:1603.07810}, 2016.

\bibitem[Xiao et~al.(2017)Xiao, Rasul, and Vollgraf]{xiao2017fashion}
Han Xiao, Kashif Rasul, and Roland Vollgraf.
\newblock Fashion-mnist: a novel image dataset for benchmarking machine
  learning algorithms, 2017.

\bibitem[Xu et~al.(2017)Xu, Sun, Deng, and Tan]{xu2017variational}
Weidi Xu, Haoze Sun, Chao Deng, and Ying Tan.
\newblock Variational autoencoder for semi-supervised text classification.
\newblock In \emph{AAAI}, pages 3358--3364, 2017.

\bibitem[Zhao et~al.(2017)Zhao, Song, and Ermon]{zhao2017infovae}
Shengjia Zhao, Jiaming Song, and Stefano Ermon.
\newblock {InfoVAE}: Information maximizing variational autoencoders.
\newblock \emph{arXiv preprint arXiv:1706.02262}, 2017.

\end{thebibliography}
